%% file: arXivFull.tex
\documentclass{article}

\usepackage{amssymb}
\usepackage{amsmath}
\usepackage{wrapfig}
\usepackage{makecell}  
\usepackage{multirow}  
\usepackage{colortbl}  
\usepackage{xcolor}  
\usepackage{graphicx}
\usepackage{pifont}
\usepackage{enumitem}

\usepackage[preprint]{corl_2026} 

\title{Dexterity-BEV: Aligning 3D World and Actions for Generalizable Robot Policies Learning}  

\author{
  \textbf{Huayi Zhou}$^{*\;1,2}$ \quad
  \textbf{Wei Gao}$^{*\;1}$ \quad
  \textbf{Dekun Lu}$^{1}$ \quad
  \textbf{Ruiji Liu}$^{1}$ \quad
  \textbf{Zhanqi Zhang}$^{1}$ \\
  \textbf{Ziyang Zhang}$^{1}$ \quad 
  \textbf{Jian Chen}$^{1}$ \quad
  \textbf{Wenlve Zhou}$^{1}$ \quad
  \textbf{Sheng Xu}$^{2}$ \quad
  \textbf{Shumin Li}$^{1}$ \\
  \textbf{Kangyi Guo}$^{1}$ \quad 
  \textbf{Shichen Xu}$^{1}$ \quad 
  \textbf{Zixin Huang}$^{1}$ \quad 
  \textbf{Yongyi Su}$^{1}$ \quad 
  \textbf{Kui Jia}$^{\ddag\;1,2}$ \\
  $^{1}$DexForce Technology \quad $^{2}$The Chinese University of Hong Kong, Shenzhen\\
  $^*$Equal Contribution \quad $^{\ddag}$Corresponding Author
}

\begin{document}
\maketitle


\vspace{-20pt}
\begin{abstract}
End-to-end manipulation policies, combined with web-scale pretrained Vision-Language Models (VLMs), show the promise for generalizable and dexterous robotic manipulation. However, they inherit two key limitations from 2D foundation models: 1) the reliance on 2D RGB inputs that ignores the intrinsically 3D nature of manipulation; and 2) the lack of spatial 3D alignment between input-output spaces as well as across diverse robot embodiments, camera setups, and trajectory datasets. In this paper, we present a series of contributions to address these issues. First, we introduce \emph{aligned vertex map} and \emph{vertex spectrum} — a pixel-wise 3D representation that elevates 2D visual inputs to 3D, using camera calibration and optional depth. This novel input representation marries 3D awareness with the generalization of 2D large VLMs. Then, we propose to align the inputs and outputs of manipulation policies by expressing per-pixel 3D information of each camera view and robot actions to a shared coordinate. Based on this, we designate a canonical \emph{Bird's-Eye-View (BEV) alignment frame} and innovatively propose to construct BEV images, producing a view-invariant representation robust to camera pose variations. To enable training and evaluation at scale, we develop a comprehensive data processing pipeline to perform such alignments; we also introduce a novel temporal alignment scheme for trajectories across diverse robots, human operators, and datasets. These contributions collectively mitigate input and output spatial-temporal misalignments, improving the consistency and generalization for real-world manipulation. Pretrained checkpoint, source code and data processing pipeline are available in \url{https://hnuzhy.github.io/projects/Dex-BEV}.
\end{abstract}



\input{tex/introduction}

\input{tex/relatedworksCompact}

\input{tex/methodology-zhy}

\input{tex/experiments}

\input{tex/conclusion}


\clearpage

\acknowledgments{This work was funded by the Key-Area Research and Development Program of Guangdong Province, China under Grant 2024B0101040004, and the Shenzhen Science and Technology Program under Grant KJZD20240903104008012 and ZDCY20250901113000001. 

Beyond that, this work was supported by the major leadership and directional guidance of Kui Jia. We sincerely thank all the contributors for their dedication: co-first authors Huayi Zhou and Wei Gao conceptualized the framework and drafted the manuscript, with Huayi Zhou conducting Agilex real-world experiments, and Wei Gao leading the simulation benchmarks, real-world deployment, and core data infrastructure; Dekun Lu and Jian Chen assisted with the data infrastructure and hardware testing; Ruiji Liu, Zhanqi Zhang, and Ziyang Zhang managed the real-robot evaluations on the A1 semi-humanoid and W1 humanoid configurations; Wenlve Zhou, Sheng Xu, and Yongyi Su contributed to text polishing and technical discussions; and Shumin Li, Kangyi Guo, Shichen Xu, and Zixin Huang supported the large-scale real-world teleoperation data collection.}


{\small  
\bibliography{example}  
}

\clearpage

\appendix

This appendix provides supplementary materials to support and expand upon the core methodologies, architectural implementations, and empirical findings presented in the main text. 
To ensure completeness, reproducibility, and rigorous academic transparency, the remainder of this document is structured into four sequential sections:
Sec.~\ref{appA} delineates the comprehensive operational definitions, hardware specifications, environmental layouts, and precise data collection protocols for all five complex, long-horizon real-world dual-arm manipulation tasks.
Sec.~\ref{appB} expands upon the mathematical formulations, structural nuances of the vertex map/spectrum encoding, and algorithmic implementations of the spatial-temporal alignment data pipeline. 
Sec.~\ref{appC} presents exhaustive quantitative performance metrics, complete baseline comparisons, extensive simulation ablations, and additional qualitative keyframe breakdowns across diverse out-of-distribution evaluation scenarios.
Sec.~\ref{appD} provides a candid, in-depth critical analysis of our framework's boundaries, ongoing technical challenges such as its reliance on explicit camera calibration, and strategic research directions for scaling BEV representation learning within embodied intelligence.

\input{tex/suppTaskSetups}

\input{tex/suppModelDetails}

\input{tex/suppExpResults}

\input{tex/suppDiscussions}


\end{document}

%% file: tex/introduction.tex
\section{Introduction}

End-to-end manipulation policies~\cite{chi2025diffusion, levine2016end, zhu2018reinforcement} offer significant potential for enabling embodied agents to understand and interact with the world. 
The success of Large Language Models (LLMs)~\cite{achiam2023gpt, bai2023qwen, touvron2023llama}, Vision-Language Models (VLMs)~\cite{wang2024qwen2vl, liu2023visual, liu2024improved} and (video) World-Models \cite{maes2026leworldmodel, gao2025seedance, team2025kling} has injected new inspiration into manipulation research.
Benefits from web-scale pretraining, these foundation models demonstrate promising zero-shot generalization.
Consequently, researchers aim to imbue robots with similar generalization capability to build robotics foundation models.

\begin{figure}
	\begin{center}
    \includegraphics[width=\linewidth]{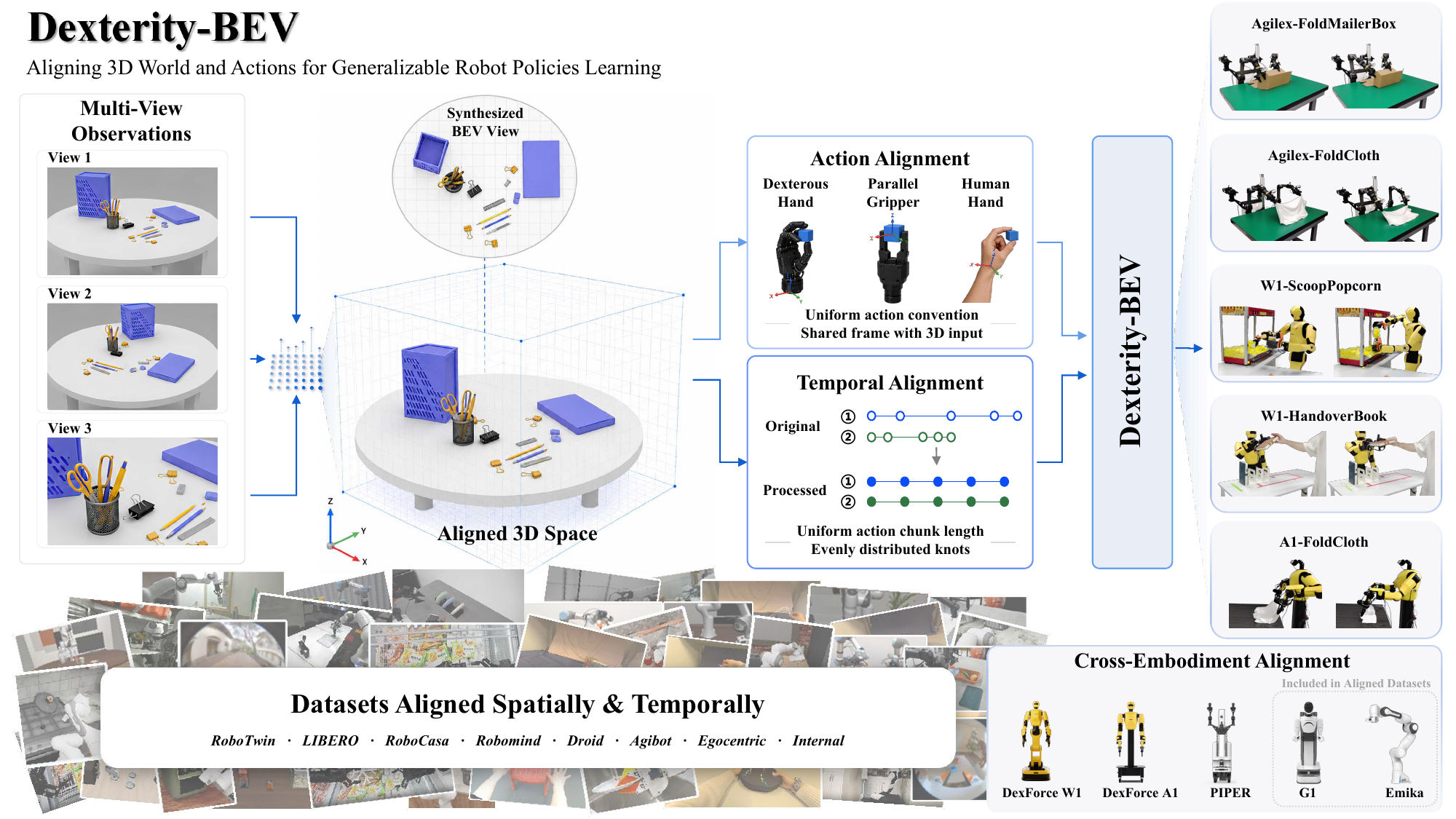}
    \vspace{-18pt}
	\caption{We introduce Dexterity-BEV (Dex-BEV), a series of technical and systematic contributions for manipulation policy learning that generalizes among different embodiments, camera views and datasets. 
    In particular, we introduce 3D input representations 
    that easily integrated with pretrained 2D VLMs; spatial alignment between multi-view cameras \& robot actions; and temporal alignment between trajectories from different robots and/or tele-operators.
    These concepts lead to a comprehensive data processing pipeline and trajectory datasets aligned spatially and temporally.
    } 
    \label{fig:teaser}
    \vspace{-15pt}
	\end{center}
\end{figure}

With this motivation, researchers are increasingly exploring Vision-Language-Action (VLAs)~\cite{kim2025openvla, black2025pi0, liu2025rdt, brohan2023rt, zitkovich2023rt, belkhale2024rt}. Some contributions augment VLAs with future video stream prediction, thus leading to World-Action Models (WAMs) \cite{ali2025world,li2026causal,gao2026dreamdojo,team2026gigabrain}. These models are usually derived from pretrained 2D VLMs and further trained on manipulation datasets, typically consisting of corresponded RGB frames, robot/human action trajectories and task instructions. 
Many dexterous manipulation behaviors challenging for traditional modular perception-planning-action pipelines, automatically emerge from these models~\cite{chi2025diffusion, kim2025openvla, black2025pi0}.

These VLAs/WAMs inherit strong capability from VLMs in terms of visual perception and textual understanding. However, VLMs are typically trained on two-dimensional (2D) RGB image and video inputs, depite robotic manipulation is intrinsically three-dimensional (3D). 
%
%
As a result, most existing endeavors~\cite{kim2025openvla, black2025pi0, liu2025rdt, brohan2023rt, zitkovich2023rt, belkhale2024rt} lack explicit 3D information from camera input, such as camera calibration results (intrinsic and extrinsic matrices) and depth images.
Consequently, several other contributions \cite{zhen20243d, li20253ds, sun2025geovla, zhang2026from, fan2026any3d} have explored alternative 3D inputs, such as point clouds and voxels.
However, datasets with these 3D representations are not yet comparable to the 2D counterparts in terms of scale and diversity, which limits the generalization of pretrained 3D VLMs/encoders, and consequently the capability of derived manipulation polices.

Moreover, the output space of existing VLAs/WAMs is typically \textit{not aligned} in terms of: 1) misalignment with (2D) input observations; and 2) misalignment across different embodiments (robots), datasets and manipulation scenarios. In particular, existing VLAs usually produce joint angles or end-effector (EE) poses as output.
Joint angles depend on robot types, thus joint trajectories to accomplish the same manipulation task can be vastly different for two types of robots. 
On the other hand, the EE pose values depend on robot types, frame conventions and designations of the ``world'' frame (in which EE poses are expressed).
For instance, the ``world'' frame depends on the table setup in the LIBERO~\cite{liu2023libero} dataset; many bi-arm manipulators (e.g., CobotMagic) express left/right EE poses in the base frames of left/right sub-arms, thus these EE pose values cannot reflect the base offset between two sub-arms.
These spatial misalignment in joint or EE space causes additional (sometimes unnecessary) variations on action trajectory distribution that end-to-end models must overcome.
In addition to 3D spatial misalignment, robot trajectory instances for a task might take different amounts of time, due to the variations in robot hardware setup and human tele-operation. This \textit{temporal misalignment} imply that the policy must address different geometric ``length'' of action chunks (explained in Subsec.~\ref{subsec:preliminary} and Subsec.~\ref{subsec:dataproc}).
All these misalignments challenge the expressiveness and generalization of policy learning.

In this paper, we make a series of technical and system contributions to mitigate these limitations. 
%
In particular,
\textbf{1)} we introduce \emph{aligned vertex map} and \emph{vertex spectrum} formation, previously used in other fields such as 3D reconstruction~\cite{wang2025vggt, lin2026depth} and autonomous driving~\cite{liu2022petr, liu2023petrv2}, into VLAs as input space.
This formulation elevates 2D-centric model inputs to 3D by providing per-pixel 3D information, exploiting camera calibrations and optional depth images. Thus, we aim to combine the benefit of 3D input space with the generalization of vision and language foundation models, pretrained on web-scale 2D datasets.
Moreover, \textbf{2)} we propose to align multi-view observations and output actions, by expressing per-pixel 3D information of each camera view, robot proprioceptive measurements and actions to a shared coordinate, thanks to camera extrinsic parameters.
Base on these formulations, \textbf{3)} we propose to designate \emph{BEV frames} (refer to Sub.~\ref{subsec:bev} for more details) as the alignment frame, and innovatively construct BEV images that are less-variant to different camera setups and change in camera view points, inspired by contributions in autonomous driving~\cite{liu2022petr, liu2023petrv2}.
To facilitate the training, evaluation and deployment of our models, we devise a comprehensive data processing pipeline with the following distinctions. Systematically, \textbf{4)} we implement 3D spatial alignment for both internal and public datasets, by combining manual operations (assisted by a customized GUI application), rule-based algorithms and vision foundation models. In addition to spatial alignment, \textbf{5)} we propose to align different trajectories temporally among different robots, tele-operators and datasets.
These contributions constitute the unified Dexterity-BEV (Dex-BEV) architecture and training receipt.
%

While these contributions above are generally applicable to both VLAs and WAMs, in this paper we focus on VLAs as the instantiated ones and defer WAMs, or the prediction of explicit future (3D) state, to a later study. Simulated and real-world experiments show that Dexterity-BEV achieves significant performance improvements given variations of camera views, robot base poses, and/or manipulation scenarios. We will make the code and data pipeline publicly available.

%% file: tex/relatedworksCompact.tex
\vspace{-5pt}
\section{Related Works}
\label{sec:related}
\vspace{-5pt}

\textbf{VLAs and WAMs.} 
The scaling of diverse robotic demonstrations pre-training has rapidly advanced VLA models. Pioneering works such as~\cite{brohan2023rt, zitkovich2023rt, kim2025openvla} validated the efficacy of VLA models derived from 2D VLMs. 
Efficient teleoperation systems like ALOHA~\cite{zhao2023learning, fu2025mobile} enabled large-scale dataset collection and spurred various VLA datasets~\cite{o2024open, khazatsky2024droid, wu2025robomind}.
%
%
%
Many contributions~\cite{team2024octo, zheng2026xvla, black2025pi0, black2025pi05} explore VLA models with different architectures, learning algorithms and auxiliary tasks. One prominent example is future video generation in WAMs~\cite{ali2025world, ye2026world, gao2026dreamdojo, li2026causal, team2026gigabrain, maes2026leworldmodel}.
%
However, these models rely predominantly on 2D image backbones. The lack of 3D input might lead to performance degradation in terms of precision and robustness to unseen camera view points.

\textbf{3D Representations in VLAs/WAMs.}
Consequently, many contributions~\cite{qu2025spatialvla, yuan2025depthvla, li20253ds, deng2025stereovla} attempt to incorporate various form of 3D input into VLAs/WAMs. 
It is straightforward to use point cloud, voxel grid, and 3D Gaussian Splatting~\cite{li2026pointvla,sun2025geovla, yu2025artgs, li20253ds} as input. However, these pure 3D representations cannot benefit from VLM backbones pretrained on web-scale 2D image and video datasets.
Another branch of contributions~\cite{deng2025stereovla, yuan2025depthvla, qu2025spatialvla} fuse 3D information into 2D VLM backbones, and our method falls into this category.
Existing methods, based on depth image~\cite{yuan2025depthvla}, stereo~\cite{deng2025stereovla} or camera-frame vertex map~\cite{li2026spatial}, typically process each camera view independently; therefore, correlation information between multiple camera views (e.g., head and wrist cameras) is not provided directly to the models. Instead, we propose to provide this information by expressing all vertex maps/spectrums in a shared BEV frame. 
This idea is extended to achieve alignment between multi-view observations, robot proprioception, and action trajectories. 
The proposed BEV image is inspired by BridgeVLA~\cite{li2026bridgevla} and autonomous driving contributions~\cite{chen2020dsgn, reading2021categorical, liu2022petr, li2022bevformer}.
Compared with~\cite{li2026bridgevla}, we further augment the RGB BEV image with a pixel-aligned vertex map. Then, an alternative network architecture and training receipt are used with emphasis on reactive manipulation tasks (e.g, cloth folding), which might be challenging for the classical motion planner in~\cite{li2026bridgevla}.

%% file: tex/methodology-zhy.tex
\vspace{-5pt}
\section{Methodology}
\label{sec:method}
\vspace{-5pt}


Dex-BEV elevates 2D-centric models into a spatially aligned 3D-aware representation for both observations and actions. This section is organized as follows: Subsec.~\ref{subsec:preliminary} provides preliminaries about VLM and VLA. Subsec.~\ref{subsec:vertex_map} details our \textit{Aligned Vertex Map Formulation} for projecting pixel features into a shared 3D frame. Subsec.~\ref{subsec:bev} extends our formulation with \textit{BEV Frame, BEV Image Construction and Vertex Spectrum}. 
Finally, Subsec.~\ref{subsec:dataproc} presents the \textit{Data Processing Pipeline} for 3D spatial standardization and temporal trajectory alignment.

\subsection{Preliminary}
\label{subsec:preliminary}

Most VLAs are derived from pretrained VLMs, which extract visual-textual representations from 2D images and instructions. Given an RGB image $\mathbf{I}_{t,i}\!\in\!\mathbb{R}^{H \times W \times 3}$ from the $i$-th camera at step $t$ and an instruction $\mathcal{L}$, the VLM extracts visual tokens $\mathbf{F}_{t,i}\!=\!\mathsf{Enc}_{vis}(\mathbf{I}_{t,i})$ and language tokens $\mathbf{E}_{lang}\!=\!\mathsf{Enc}_{lang}(\mathcal{L})$. Multi-view visual tokens are aggregated into $\tilde{\mathbf{F}}_{t}$, which is further fused into contextual embedding $\mathbf{c}_t\!=\!\mathcal{F}_{\theta}(\tilde{\mathbf{F}}_{t}, \mathbf{E}_{lang})$.

VLAs predict robot actions from multimodal state $\mathcal{X}_t = \{ \{ (\mathbf{O}_{t,i}, \mathbf{K}_i, \mathbf{T}_{t, i}) \}_{i=1}^N, \mathcal{L}, \mathbf{s}_{t} \}$ at each step $t$.
$\mathbf{O}_{t,i}$ contains an RGB image $\mathbf{I}_{t,i}$ and an optional, pixel-aligned depth map $\mathbf{D}_{t,i} \in \mathbb{R}^{H \times W}$, where $N$ is the number of cameras. The matrices $\mathbf{K}_i \in \mathbb{R}^{3 \times 3}$ and $\mathbf{T}_{t, i} \in SE(3)$ denote camera intrinsics and extrinsics, respectively. Given the input $\mathcal{X}_t$, the VLA policy predicts a chunk of $M$ future actions $\{\mathbf{A}_{t+m}\}_{m=1}^M$. Recent VLA models condition an action decoder on the VLM embedding $\mathbf{c}_t$ using Flow Matching (FM)~\cite{lipman2023flow, black2025pi0, black2025pi05} to model precise action distributions. FM trains a vector field $\mathbf{v}_{\theta}(\mathbf{a}_{\sigma}, \sigma, \mathbf{c}_t)$ along a probability path $\psi_{\sigma}(\mathbf{a}) = \sigma \mathbf{a}_1 + (1 - \sigma) \mathbf{a}_0$ between Gaussian noise $\mathbf{a}_0 \sim \mathcal{N}(0, \mathbf{I})$ and ground-truth actions $\mathbf{a}_1$ by minimizing:
\begin{equation}
	\centering
	\mathcal{L}_{FM} = \mathbb{E}_{\sigma \sim \mathcal{U}[0,1], \mathbf{a}_1 \sim p_{data}, \mathbf{a}_0 \sim p_0} \left[ \| \mathbf{v}_{\theta}(\sigma \mathbf{a}_1 + (1-\sigma)\mathbf{a}_0, \sigma, \mathbf{c}_t) - (\mathbf{a}_1 - \mathbf{a}_0) \|^2 \right].
	\label{eqnA}
\end{equation}
During inference, the action sequence is sampled via an ODE solver: $\mathbf{a}_1 = \mathbf{a}_0 + \int_0^1 \mathbf{v}_{\theta}(\mathbf{a}_{\sigma}, \sigma, \mathbf{c}_t) d\sigma$.

\subsection{Aligned Vertex Map Formulation}
\label{subsec:vertex_map}

Following Subsec.~\ref{subsec:preliminary}, the observation at step $t$ is defined as $\mathcal{X}_t\!=\!\{ \{ (\mathbf{O}_{t,i}, \mathbf{K}_i, \mathbf{T}_{t, i}) \}_{i=1}^N, \mathcal{L}, \mathbf{s}_{t} \}$. In this subsection, we assume all cameras are calibrated and depth images are available, thus the observation becomes $\mathbf{O}_{t,i} = (\mathbf{I}_{t,i}, \mathbf{D}_{t,i})$. \emph{This assumption is relaxed in Sec.~\ref{subsec:bev} to address setups without depth images on one or more camera views.} Given depth map $\mathbf{D}_{t,i}$ and intrinsics $\mathbf{K}_i$, the pixel $(u, v)$ is back-projected to obtain a 3D vertex in the $i$-th camera frame:
\begin{equation}
	\mathbf{P}_{camera\_i}(u, v) = \mathbf{K}_i^{-1} [u, v, 1]^T \mathbf{D}_{t,i}(u, v),
	\label{equ:d2vm}
\end{equation}
\noindent where $\mathbf{P}_{camera\_i}$ is a vertex map, and the time subscript $t$ is omitted for clarity. The 2D pixel structure of $\mathbf{P}_{camera\_i}$ enables easy integration into 2D VLMs. Prior methods like SpatialVLA~\cite{qu2025spatialvla} directly leverage this local map to formulate 3D positional embeddings for visual features:
\begin{equation} \label{equ:spatialvla}
    \mathbf{F_{combined\_i}} = \mathbf{F_{img\_i}} + \mathbf{F_{3d\_i}} = \mathsf{Enc}_{vis}(\mathbf{I}_{t,i}) + \mathsf{Enc}_{3d}(\mathbf{P}_{camera\_i}).
\end{equation}
However, local vertex maps $\mathbf{P}_{camera\_i}$ lack geometric correlation across distinct viewpoints. A single physical 3D point observed across multiple views will yield highly divergent values due to differing camera extrinsics $\mathbf{T}_{t, i}$ and $\mathbf{T}_{t, j}$. 
Inspired by contributions~\cite{wang2025vggt, lin2026depth} in 3D reconstruction,
we propose to transform all camera-frame vertex maps into a shared reference frame $\mathbf{T}_{align\_t}$:
\begin{equation} \label{equ:spatialvla_aligned}
    \mathbf{F_{3d\_i}} = \mathsf{Enc}_{3d}(\mathbf{P_{aligned\_i}}) = \mathsf{Enc}_{3d}( \mathbf{T}_{align\_t}^{-1} \mathbf{T_{t,i}} \mathbf{P}_{camera\_i} ).
\end{equation}
\noindent This step ensures that $\mathbf{P_{aligned\_i}}$ maintains global spatial consistency in 3D while remaining pixel-aligned with RGB image $\mathbf{I}_{t,i}$. Crucially, the robot proprioception $\mathbf{s}_{t, i}$ and target actions $\mathbf{A}_{t}$ are also represented as $SE(3)$ poses expressed in this shared $\mathbf{T}_{align\_t}$ frame. \emph{Combined with unified 3D frame conventions (detailed in Subsec.~\ref{subsec:dataproc}), the entire perception-action loop is tightly integrated within an embodiment-agnostic 3D workspace.}

The $\mathbf{T}_{align\_t}$ is typically the first camera view in 3D reconstruction. In this paper, we instantiate $\mathbf{T}_{align\_t}$ as a canonical Bird's-Eye View (\textbf{BEV}) frame and construct additional BEV images, as detailed in Subsec.~\ref{subsec:bev}.

\begin{figure}
	\begin{center}
    \includegraphics[width=\linewidth]{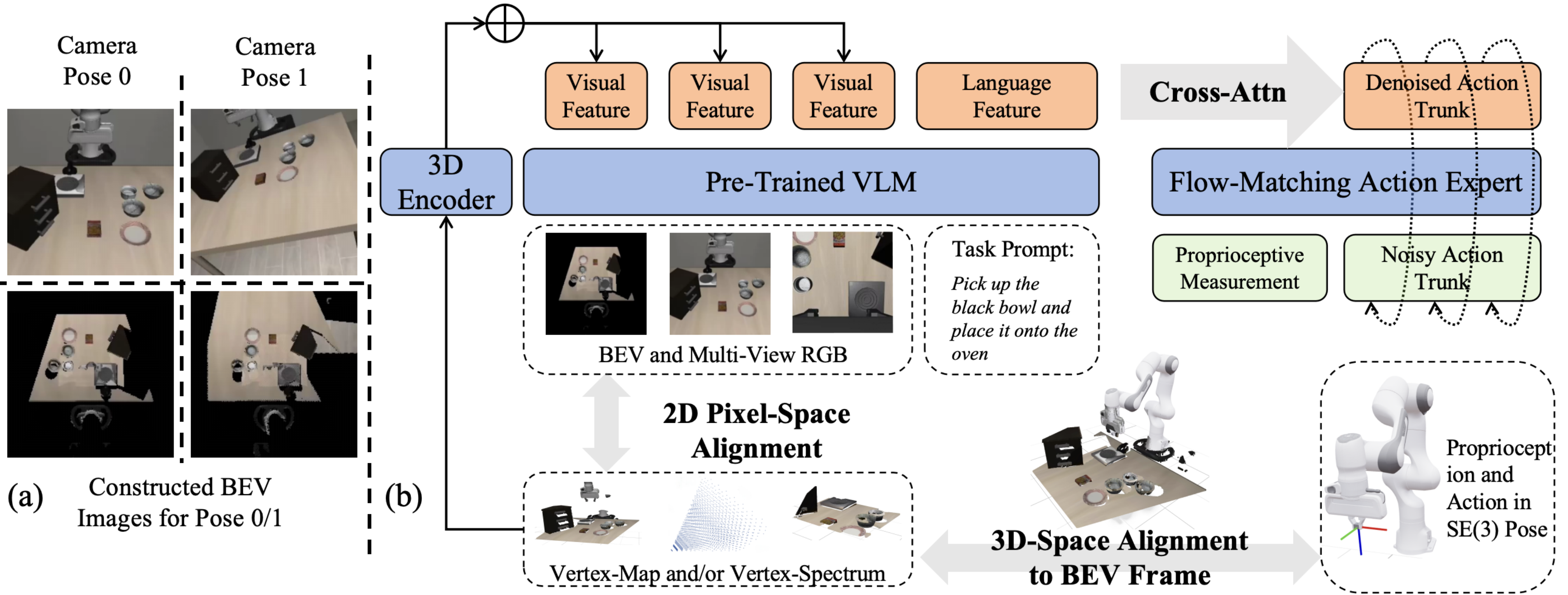}
	\vspace{-15pt}
	\caption{(a) We propose to construct BEV images and associated vertex maps towards invariance to different camera view points. Note that the synthesized BEV images for two vastly different camera poses are very similar to each other, and objects are located at almost identical pixel locations in BEV images. (b) An overview of Dex-BEV architecture. Please refer Sec.~\ref{subsec:bev} for a detailed explanation. } 
    \label{fig:model}
	\vspace{-10pt}
	\end{center}
\end{figure}

\subsection{Several Extensions and Network Architecture}
\label{subsec:bev}

\textbf{BEV Frame and BEV Image Construction.}
To minimize input variations caused by heterogeneous robotic embodiments and diverse camera setups, we formalize the shared alignment frame $\mathbf{T}_{align\_t}$ as a canonical Bird's-Eye View (BEV) reference frame. Following the conventions in autonomous driving~\cite{liu2022petr, liu2023petrv2} (``lidar frame'') and BridgeVLA~\cite{li2026bridgevla}, we instantiate $\mathbf{T}_{align\_t}$ as either: 1) the robot base frame; or 2) the bottom-center of a 3D cubic region-of-interest (RoI) surrounding the table-top workspace, if the scenario is a table-top manipulation. 

For the designated BEV frame, we construct a synthetic BEV image inspired by contributions~\cite{chen2020dsgn, reading2021categorical, liu2022petr, li2022bevformer} in autonomous driving. This BEV image is constructed by a top-down orthographic projection of the aggregated colored point clouds from all cameras. Alongside this projection, we compute a corresponding pixel-wise 3D vertex map for the BEV image. An illustration is shown in Fig.~\ref{fig:model} (a), the BEV images provide a viewpoint-invariant geometric input space for policy learning.

\textbf{Vertex Spectrum to Address Optional Depth Observation.}
%
%
Inspired by PETR in autonomous driving~\cite{liu2022petr, liu2023petrv2}, we propose generating a \textit{vertex spectrum} for RGB-only views, in order to accommodate platforms without depth sensors.
For a pixel $\mathbf{p}\!=\![u, v, 1]^T$ in the $i$-th camera view, we sample $M$ discrete depth hypotheses $d_j$ using a linear-increasing discretization (LID) \cite{reading2021categorical}:
\begin{equation}
	d_j = d_{min} + (d_{max} - d_{min}) \cdot \frac{j(j+1)}{M(M+1)},
	\label{eqnB}
\end{equation}
\noindent where $[d_{min}, d_{max}]$ represents the operational depth range. Each pixel-depth pair is back-projected and transformed via the extrinsic matrix $\mathbf{T}_{t,i}$ into the aligned BEV frame, yielding a volumetric coordinate grid $\mathcal{G}_{u,v}\!\in\!\mathbb{R}^{M \times 3}$. This grid is then processed by a lightweight encoder to formulate a 2D positional embedding that is element-wise added to the corresponding RGB features.

\textbf{Overall Architecture.}
As illustrated in Fig.~\ref{fig:model} (b), the overall architecture ingests these fused multi-view tokens, the synthetic BEV features, 3D vertex maps \& vertex spectrum, and the language instruction into a VLM backbone. The extracted multi-modal representations are then processed by a flow-matching action expert~\cite{black2025pi0, black2025pi05} to model the target action distribution. Crucially, both proprioceptive measurement and output action for robots are parameterized as $SE(3)$ poses expressed within the unified BEV frame. Thus, the multi-view input and action output are aligned in 3D space, and unified convention can be applied across different embodiments and datasets.

\subsection{Data Alignment Processing Pipeline}
\label{subsec:dataproc}

To facilitate robust training, evaluation, and cross-platform deployment, we implement a comprehensive pipeline for 3D spatial and temporal alignments across heterogeneous datasets.

\begin{figure}
	\begin{center}
    \includegraphics[width=1.00\linewidth]{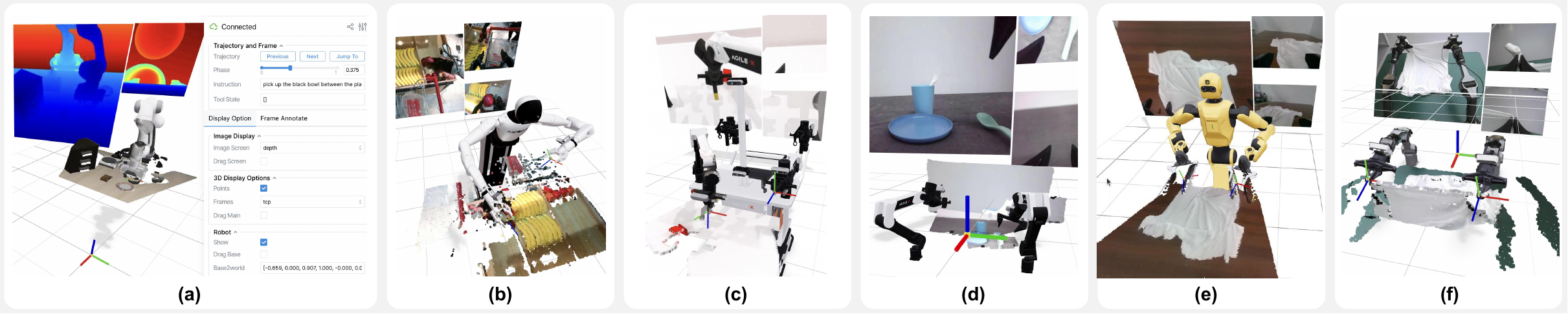}
	\vspace{-15pt}
	\caption{\textbf{3D spatial alignment in our data processing pipeline.} (a) We develop a customized GUI application for 3D alignment and visualization, as explained in Subsec.~\ref{subsec:dataproc}. In (a-f), we show the 3D alignment of representative public and internal datasets, including (a) LIBERO~\cite{liu2023libero}, (b) Agibot-Alpha/Beta~\cite{bu2025agibot}, (c) RoboTwin 2.0~\cite{mu2025robotwin}, (d) RoboMind 2.0~\cite{wu2025robomind} and our internal datasets (e-f). We also apply an unified TCP frame convention, as shown in these figures. } 
    \label{fig:dataproc}
	\vspace{-15pt}
	\end{center}
\end{figure}

\textbf{3D Spatial Alignment.}
As shown in Fig.~\ref{fig:dataproc}, for each dataset, camera intrinsics and extrinsics are unified into standard OpenCV formats by combining manual 3D GUI matching, iterative closest point (ICP) registration, and data-driven estimators like DepthAnything V3 \cite{lin2026depth}. For trajectories lacking active depth measurements, missing channels are re-generated by replaying actions in simulation; for some real-world dataset (e.g., Droid), depth images can be synthesized using vision foundation models such as FoundationStereo \cite{wen2025foundationstereo}. Finally, high-quality robot URDF models are registered to the shared 3D observation space. We enforce a unified tool center point (TCP) convention across disparate embodiments, consistently anchoring parallel-jaw gripper frames at the tip of the jaws and multi-finger configurations at the wrist. These standardized kinematic chains allow us to compute unified absolute $SE(3)$ poses across all platforms using forward kinematics.

\textbf{Cross-Trajectory Temporal Alignment.}
The speed of a trajectory can depend on the robot platform and human teleoperation, which creates additional variations on robot trajectories that VLAs must overcome. On the other hand, most manipulation tasks can be regarded as \textit{quasi-static}: a slowed or accelerated (within an extent) trajectory can still accomplish the given manipulation task. 
Although this is not true for all manipulations (e.g., throwing a ball), nearly all tasks in current VLA datasets are quasi-static.
With this observation, we propose normalizing the end-effector speed to a standard value across multiple robots and VLA datasets. In other words, we re-compute the physical time for knots of robot trajectories for temporal alignment. The detailed procedure is in Appendix.

%% file: tex/experiments.tex
\vspace{-5pt}
\section{Experiments}
\label{sec:results}
\vspace{-5pt}

Our evaluation aims to demonstrate that Dex-BEV provides a superior and more interpretable framework for dexterous robotic manipulation compared to existing 2D and 3D VLA paradigms. We systematically test its efficacy across diverse simulated benchmarks \cite{liu2023libero, mu2025robotwin, chen2025robotwin} and real-world platforms, focusing on its spatial reasoning capabilities and cross-embodiment generalization. 

\vspace{-5pt}
\subsection{Evaluation on Simulation Benchmarks}
\vspace{-5pt}

We perform quantitative comparisons on the LIBERO \cite{liu2023libero} and RoboTwin-2.0 \cite{mu2025robotwin, chen2025robotwin} benchmarks. Our method is compared with two competitive VLA baselines: the $\pi_0$~\cite{black2025pi0} and X-VLA~\cite{zheng2026xvla} \footnote{We emphasize that given the complementary nature of our proposed Dex-BEV, 
similar quality results would be obtained if comparing with other representative VLAs \cite{liu2025rdt,black2025pi05,kim2025openvla}.}. 
Moreover, we conduct a 2D ablation study of the proposed method that 1) removes all 3D inputs; and 2) disables 3D alignment by expressing all SE(3) poses following the conventions of X-VLA~\cite{zheng2026xvla}. As detailed below, we use the official and modified setups to evaluate the generalization of our method with respect to different embodiments, camera viewpoints and robot/scene base poses.

We first evaluate our method on the official setup of LIBERO \cite{liu2023libero} and RoboTwin-2.0 \cite{chen2025robotwin}. These benchmarks are based on different robot platforms, the single-arm 7-DoF franka for LIBERO \cite{liu2023libero} and dual-arm 12-DoF agile-x for RoboTwin-\cite{chen2025robotwin}. The results are shown in Tab.~\ref{tabSimResultsMixed}. To highlight the generalization to different embodiments, we use one checkpoint (network weight) for both evaluation. The results for baselines are the higher one from our rollout of released checkpoints and the reported results in~\cite{zheng2026xvla}. Compared with these SOTA baselines, our method achieves roughly the same results on LIBERO~\cite{liu2023libero} and higher success rate on RoboTwin~\cite{mu2025robotwin} in Tab.~\ref{tabSimResultsMixed}, despite deploying on vastly different robot platforms. Moreover, the 2D ablation, which use the same input/output as X-VLA~\cite{zheng2026xvla}, suffers major performance drop. This highlights the effectiveness of proposed 3D inputs and alignments.

\begin{table}[]
    \centering
    \caption{\textbf{Simulation benchmark results and generalization to different embodiments.} We present the success rate for each compared method across task suites in LIBERO~\cite{liu2023libero} and RoboTwin 2.0~\cite{mu2025robotwin}. Our method achieves roughly the same results on LIBERO and higher success rate on RoboTwin compared to strong baselines, despite deploying on vastly different robot platforms. In comparison with 2D ablation, the proposed 3D inputs and alignments lead to major improvement.
    }
	\setlength{\tabcolsep}{3pt}
    \begin{tabular}{c|c|cccc|c|cc}
	\Xhline{1.2pt}
    \multirow{2}{*}{Method} & \multirow{2}{*}{\begin{tabular}[c]{@{}c@{}}Cross \\ Embodiments\end{tabular}} & \multicolumn{5}{c|}{LIBERO (Official)}    & \multicolumn{2}{c}{RoboTwin 2.0} \\
    \cline{3-9} 
    ~ & ~ & \cellcolor{gray!15} Spatial & \cellcolor{gray!15} Object & \cellcolor{gray!15} Goal & \cellcolor{gray!15} Long & \cellcolor{gray!15} Average & \cellcolor{gray!15} Clean & \cellcolor{gray!15} Randomized        \\
    \hline
    $\pi_0$~\cite{black2025pi0}     & False & 96.8 & 98.8 & 95.8 & 85.2 & 94.2 & 46.4 & 16.4 \\
    X-VLA~\cite{zheng2026xvla}      & False & 98.2 & 98.6 & 97.8 & 97.6 & 98.1 & 70.0 & 39.0 \\
    \hline
    2D Ablation                     & True  & 93.2 & 95.0 & 92.8 & 90.2 & 92.8 & 64.8 & 35.2 \\
    Dex-BEV                        & True  & 98.2 & 98.0 & 97.8 & 97.0 & 97.8 & 76.0 & 42.0 \\
	\Xhline{1.2pt}
    \end{tabular}
    \label{tabSimResultsMixed}
    \vspace{-10pt}
\end{table}

\makeatletter\def\@captype{table}\makeatother
\begin{minipage}{.65\columnwidth}\footnotesize  
	\centering
	\vspace{-5pt}
	\caption{\textbf{Modified LIBERO benchmark to evaluate generalization to camera view points and robot/scene base poses.} The proposed method achieves reasonable success rate despite significant variations on camera viewpoints and base poses of robot \& scene (everything except the robot, such as the table and objects).}
	\vspace{5pt}
	\label{tabSimResults}
	\setlength{\tabcolsep}{3pt}
	\begin{tabular}{c|cccc|c}
	\Xhline{1.2pt}
	\multirow{2}{*}{Method} & \multicolumn{5}{c}{\makecell{Modified LIBERO\\(Mutated Camera \& Scene Layout)}} \\
    \cline{2-6}
    ~ & \cellcolor{gray!15} Spatial & \cellcolor{gray!15} Object & \cellcolor{gray!15} Goal & \cellcolor{gray!15} Long & \cellcolor{gray!15} Average \\
	\hline
    X-VLA (official ckpt) & $<$10 & $<$10 & $<$10 & $<$10 & $<$10 \\
    2D Ablation & $<$10 & $<$10 & $<$10 & $<$10 & $<$10 \\
    Dex-BEV & 92.8 & 89.4 & 91.0 & 86.2 & 89.9 \\
	\Xhline{1.2pt}
	\end{tabular}
\end{minipage}
\hspace{1pt}
\makeatletter\def\@captype{figure}\makeatother
\hspace{0.0cm}\begin{minipage}{.34\columnwidth}
	\centering
	\vspace{3pt}
	\includegraphics[width=\columnwidth]{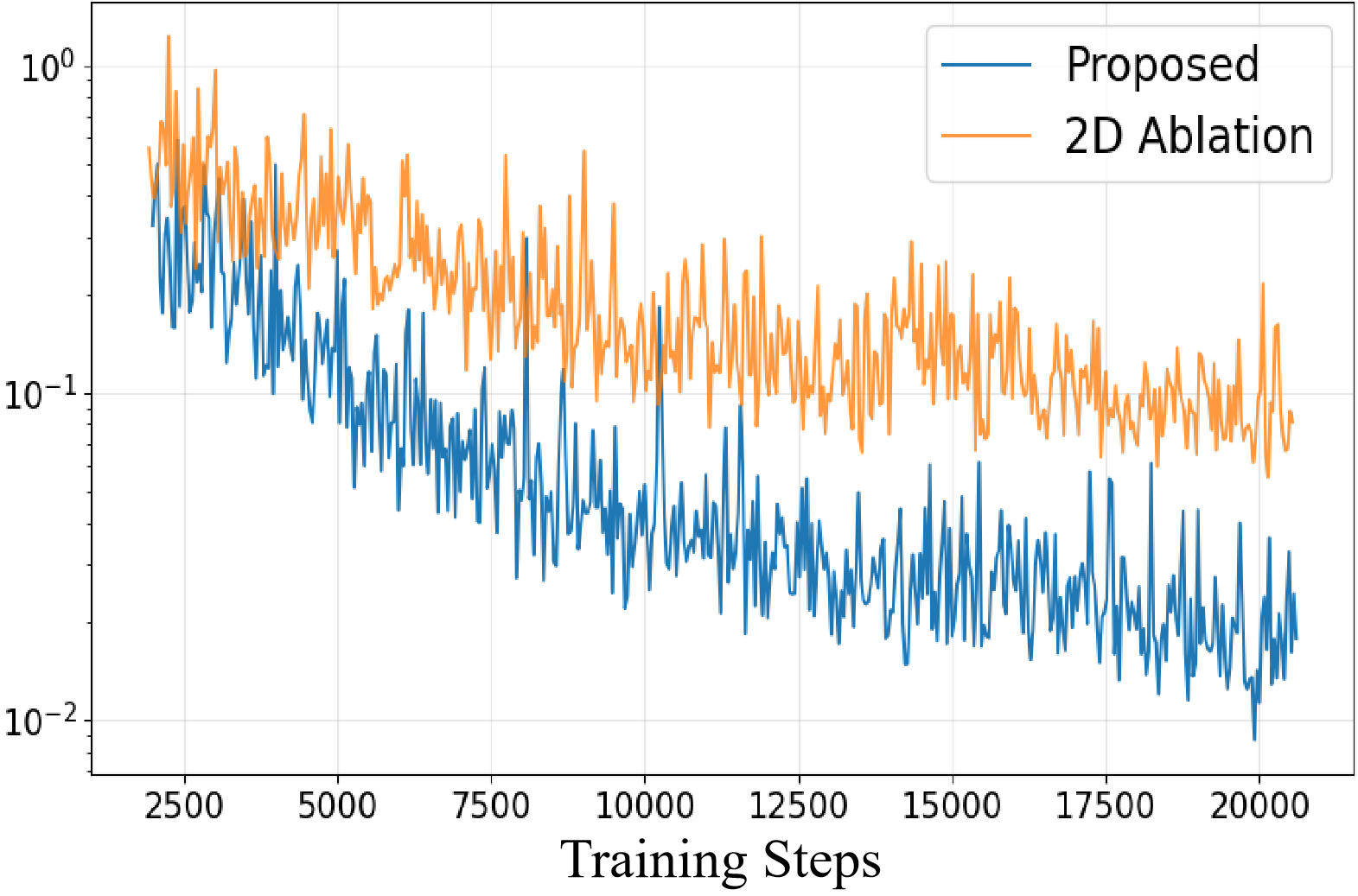}
	\vspace{-10pt}
	\caption{\textbf{Training loss comparison.} This corresponds to Tab.~\ref{tabSimResults}.} 
	\label{figSimResults}
\end{minipage}
\vspace{0.2cm}

We further conduct simulated experiments to exanimate the generalization to different camera view points and environment layouts. To achieve this, we modify the setup on the LIBERO~\cite{liu2023libero} datasets and platforms. In particular, for each trajectory we randomly modify the third-view camera pose by placing it at different distance and rotating it relative to the world-$z$ axis, the optical axis and the tilting angle. 
Moreover, we apply local 6-DoF random perturbation to the base pose of the robot and scene (everything except the robot, such as the table and objects) for each trajectory.
During the re-generation of LIBERO demonstration trajectory, we first move the robot end-effector to compensate the movement of robot and scene base pose.

The simulation results are presented in Tab.~\ref{tabSimResults}. The official X-VLA~\cite{zheng2026xvla} checkpoint and 2D ablation cannot address strong perturbation of camera poses and scene layouts above. On the other hand, our method achieves a reasonable success rate in this evaluation, benefiting from the representation and alignment of the 3D input. Fig.~\ref{figSimResults} compares the training dynamics of our method and 2D ablation. The 2D baseline cannot adequately adsorb the pose variations in the training data.

\subsection{Evaluation on Real-World Platforms}
\label{subsec:real_world_eval}

To validate the practical utility, robustness, and physical precision of Dex-BEV, we deploy our framework across four distinct dual-arm hardware setups: an Agilex bimanual platform, two DexForce wheeled-humanoid robots equipped with two dexterous hands (W1*) or parallel grippers (W1), and a DexForce A1 semi-humanoid robot. Our real-world evaluation comprises five long-horizon tasks that involve intricate bimanual coordination and interactions with deformable, articulated, or granular objects: (1) \texttt{Fold Mailer Box} and (2) \texttt{Fold Cloth} on the Agilex platform; (3) \texttt{Scoop Popcorn} and (4) \texttt{Handover Book} on the W1 humanoid; and (5) \texttt{Fold Cloth} on the A1 semi-humanoid with .
For these different robotic embodiments and corresponding task rollout examples, please refer Fig.~\ref{fig:teaser} right and Fig.~\ref{figResults} for more details.
These scenarios present high-dimensional joint synchronization, and multi-contact dynamics, making them inherently challenging for 2D-aware policies. We baseline our framework against strong competitors, including $\pi_0$~\cite{black2025pi0} and X-VLA~\cite{zheng2026xvla}. As quantitatively shown in Tab.~\ref{tabResults}, Dex-BEV demonstrates a stable execution profile and commands a significant success rate advantage over all baselines, establishing a new state-of-the-art for physical dual-arm dexterity. 

\makeatletter\def\@captype{table}\makeatother
\begin{minipage}{.31\columnwidth}\scriptsize  
	\centering
	\vspace{-5pt}
	\caption{Quantitative comparison results of real-robot experiments (reporting average success rates across 30 trails).}
	\vspace{3pt}
	\label{tabResults}
	\setlength{\tabcolsep}{3pt}
	\begin{tabular}{c|c}
	\Xhline{1.2pt}\rowcolor{gray!15}
	Task 1 (Agilex) & \texttt{Fold Mailer Box} \\
	\hline
	  $\pi_0$ \cite{black2025pi0} & 13/30 (43.3\%) \\  
    X-VLA \cite{zheng2026xvla} & 17/30 (56.7\%) \\  
    Dex-BEV & \textbf{23/30 (76.7\%)} \\  
    \Xhline{0.9pt}\rowcolor{gray!15}
	Task 2 (Agilex) & \texttt{Fold Cloth} \\
	\hline
	  $\pi_0$ \cite{black2025pi0} & 20/30 (66.7\%) \\  
    X-VLA \cite{zheng2026xvla} & 24/30 (80.0\%) \\  
    Dex-BEV & \textbf{28/30 (93.3\%)} \\  
    \Xhline{0.9pt}\rowcolor{gray!15}
	Task 3 (W1*) & \texttt{Scoop Popcorn} \\
	\hline
	  $\pi_0$ \cite{black2025pi0} & 18/30 (60.0\%) \\  
    X-VLA \cite{zheng2026xvla} & 21/30 (70.0\%) \\  
    Dex-BEV & \textbf{26/30 (86.7\%)} \\  
    \Xhline{0.9pt}\rowcolor{gray!15}
	Task 4 (W1) & \texttt{Handover Book} \\
	\hline
	  $\pi_0$ \cite{black2025pi0} & 12/30 (40.0\%) \\  
    X-VLA \cite{zheng2026xvla} & 21/30 (70.0\%) \\  
    Dex-BEV & \textbf{28/30 (93.3\%)} \\  
    \Xhline{0.9pt}\rowcolor{gray!15}
	Task 5 (A1) & \texttt{Fold Cloth} \\
	\hline
	  $\pi_0$ \cite{black2025pi0} & 19/30 (63.3\%) \\  
    X-VLA \cite{zheng2026xvla} & 23/30 (76.7\%) \\  
    Dex-BEV & \textbf{29/30 (96.7\%)} \\  
	\Xhline{1.2pt}
	\end{tabular}
\end{minipage}
\hspace{1pt}
\makeatletter\def\@captype{figure}\makeatother
\hspace{0.0cm}\begin{minipage}{.67\columnwidth}
	\centering
	\vspace{3pt}
	\includegraphics[width=\columnwidth]{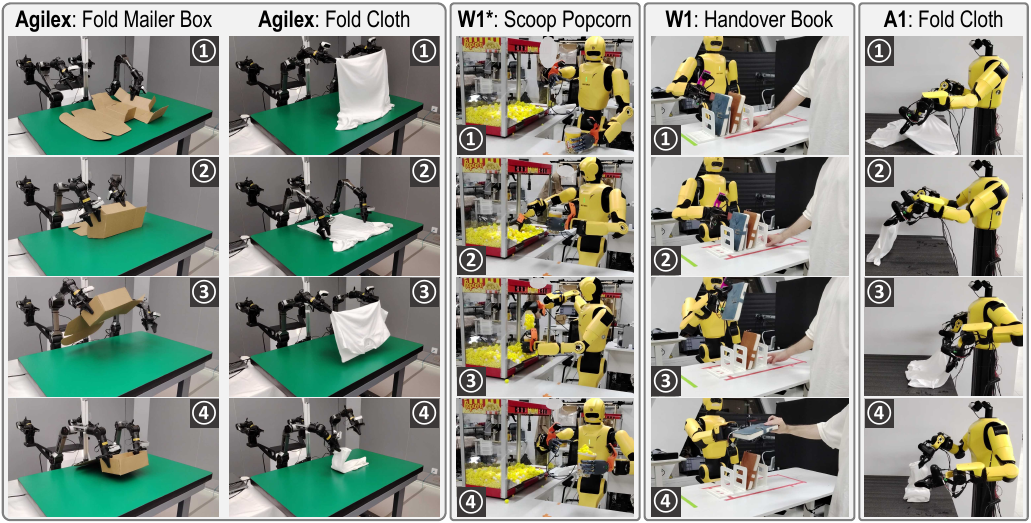}
	\vspace{-15pt}
	\caption{\textbf{Qualitative real-world rollouts across different long-horizon complex tasks.} Distinct keyframes demonstrate successful autonomous executions on diverse bimanual robotic platforms involving articulated, deformable, and granular objects (from left to right): \texttt{Fold Mailer Box} and \texttt{Fold Cloth} on Agilex, \texttt{Scoop Popcorn} and \texttt{Handover Book} on the DexForce W1 humanoid, and \texttt{Fold Cloth} on the DexForce A1 semi-humanoid.} 
	\label{figResults}
\end{minipage}
\vspace{0.3cm}

Qualitative rollout sequences, displayed in Fig.~\ref{figResults}, highlight the model's remarkable closed-loop reactivity to environmental alterations and its OOD generalization. Specifically, for the folding tasks (\texttt{Fold Mailer Box} and \texttt{Fold Cloth}), although the training demonstrations were limited to a fixed set of canonical items (e.g., white T-shirts), Dex-BEV achieves successful zero-shot adaptation to completely unseen colors, sizes, and rigidities. The \texttt{Scoop Popcorn} task demands fine-grained control over granular materials where the model must dynamically correct its trajectory despite unexpected manual displacements of the target cup. For the \texttt{Handover Book} task, Dex-BEV effortlessly handles dynamic human-in-the-loop interactions, accurately tracking a human partner's moving hand and submitting the object despite hand occlusions and unpredictable timing. To sum up, the resilience to dynamic workspace disturbances and large geometric shifts confirms that our framework models the underlying 3D spatial mechanics of a task rather than memorizing superficial 2D visual patterns. Due to page constraints, complete details regarding data collection protocols, teleoperation setups, and additional hardware specifications are expanded in the \textbf{Appendix}, with full dynamic executions provided in the \textbf{Supplementary Videos}.

%% file: tex/conclusion.tex
\section{Conclusion}
\label{sec:conclude}

This paper introduces Dexterity-BEV (Dex-BEV), a framework that establishes a unified input-output 3D alignment for generalizable and dexterous robotic manipulation. 
We bring in both vertex map and vertex spectrum as input representation for these end-to-end manipulation policies. Then, we designate the BEV frame and propose to construct BEV images, as steps towards spatial transparency and viewpoint invariance. We further propose to align trajectories temporally to mitigate the variance among different robots, tele-operators and datasets. 
Systematically, we implement a data processing pipeline that combines GUI-assisted manual operations, rule-based algorithms, and vision foundation models for spatial and temporal alignment.
Extensive experiments in simulation and real-world demonstrate the efficacy and superiority of our method.


\textbf{Limitations:} Despite these results, Dex-BEV currently relies on camera calibration, which may limit its immediate deployment in unstructured environments where extrinsic parameters are unknown. 
Future research might explore calibration-free BEV lifting through end-to-end geometric prior learning. Alternatively, advances in foundation models for 3D reconstruction~\cite{wang2025vggt,lin2026depth,wang2026vggt} can be used to obtain camera parameters, although our experience in data processing indicates these models might need more effort towards universally reliable for online, reactive robotic manipulation applications.
Scaling this architecture to more heterogeneous datasets will further solidify BEV representations as a universal and scalable interface for embodied intelligence.

%% file: tex/suppTaskSetups.tex
\section{More Details of Manipulation Tasks and Setups}\label{appA}

To provide maximum technical clarity, and hardware transparency, this section details the physical task definitions, robotic platforms, sensory configurations, and teleoperation data collection protocols for our five real-world long-horizon bimanual manipulation evaluation benchmarks.

 \begin{figure}[h]
	\begin{center}
    \includegraphics[width=1.0\linewidth]{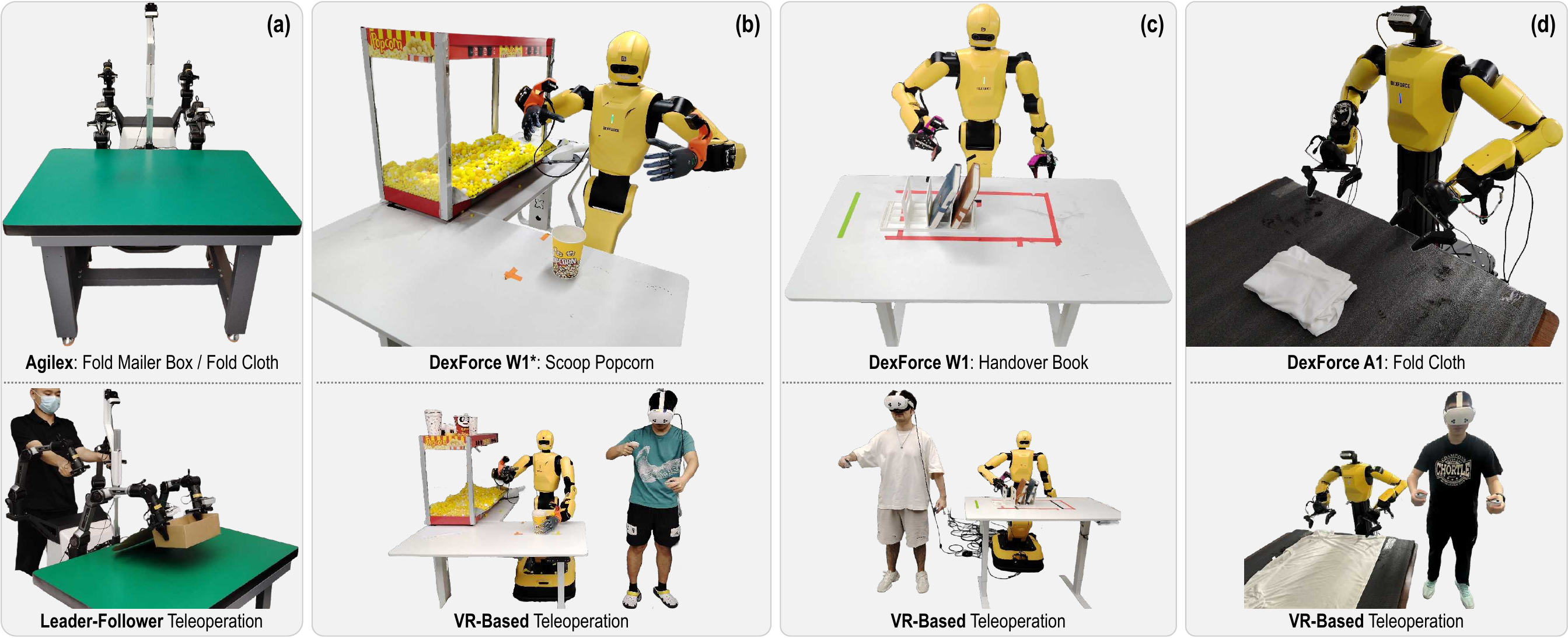}
	\vspace{-15pt}
	\caption{\textbf{Hardware, Platforms and Teleoperation Data Collection Interfaces.} From left to right: (a) the Agilex dual-arm robot platform using grippers, (b) the W1 humanoid robot platform using dexterous hands, (c) the W1 humanoid robot platform using grippers, and (d) the A1 semi-humanoid robot platform using grippers.} 
    \label{platforms}
	\vspace{-10pt}
	\end{center}
\end{figure}

\subsection{Agilex Bimanual Setup: \texttt{Fold Mailer Box} \& \texttt{Fold Cloth}}
\label{subsec:agilex_setup}

\textbf{Hardware and Sensory Configuration:} Following the standard hardware parameters specified in X-VLA \cite{zheng2026xvla}, the platform comprises two 6-DoF PiPER mechanical arms equipped with dual parallel-jaw grippers (Fig.~\ref{platforms}(a)). The vision system consists of one table-mounted central head camera and two arm-mounted wrist cameras, all instantiated via Orbbec DaBai binocular depth sensors. Although the native sensors support active depth channels, all data logging and policy inference operations utilize only the RGB streams operating at 30 FPS to ensure systemic throughput and computation efficiency. Hand-eye calibration is explicitly performed across all three views relative to the tabletop workspace boundary to facilitate the fast geometric synthesis of the Bird's-Eye-View (BEV) input frame for supporting the Dex-BEV's finetuning and deployment.

\textbf{Task \texttt{Fold Mailer Box}:} This task requires the dual-arm system to construct a structural 3D container out of an initially disassembled mailer box. Given the extreme structural complexity of folding a rigid-articulated item from scratch, the initial state is systematically simplified to prevent complete planar adherence to the workspace surface (non-prehensile grasping failure). All critical joints of the box are pre-creased, and the box is placed inside-up. No horizontal orientation constraints are enforced during deployment, which means the initial position and yaw angles are randomly mutated across the active workspace area during data collection to enrich spatial data distribution and guarantee model convergence. To counteract the out-of-distribution (OOD) structural traits of this task relative to general internet-scale pre-training datasets, we collect \textbf{1,500 teleoperated demonstration} trajectories. The duration per rollout \textit{scales non-uniformly from 30 to 45 seconds} due to variation in the pre-manipulation re-orientation steps and occasional correction maneuvers.

\textbf{Task \texttt{Fold Cloth}:} To evaluate the model's policy robustness when interacting with highly deformable objects, this task presents an unconstrained cloth folding assignment. Rather than initiating from an extreme knot condition, the initial garment state alternates between a randomized crumpled configuration or a pre-flattened placement. The goal-conditioned policy must decouple this task into two primitive capacities: active flattening followed by geometric folding. Since primitive garment manipulation layouts may exist within internet-scale pre-training distributions, we collect a compact set of \textbf{400 demonstrations} for target downstream VLA fine-tuning. The execution timeframe per trajectory \textit{spans between 50 and 75 seconds}, where the iterative smoothing and flattening phase introduces the highest variance into the sub-step execution duration.

\subsection{DexForce W1 Humanoid Setup: \texttt{Scoop Popcorn}}
\label{subsec:w1_scoop}

\textbf{Hardware and Sensory Configuration:} This task utilizes the DexForce W1 wheeled-humanoid mobile robot platform. The end-effectors are configured with dual 6-DoF five-finger BrainCo Revo-2 dexterous anthropomorphic hands (Fig.~\ref{platforms}(b)). Visual observation is captured by a centralized KingFisher CV1 binocular camera on the robot's head, augmented by two RealSense D405 binocular sensors mounted rigidly to the wrist joints. To handle the heterogeneous camera profiles efficiently, data logging frequency is standardized at 30 FPS (the action execution frequency is downsampled into 10 FPS for improving efficiency).

\textbf{Task Definition and Constraints:} Teleoperation data collection is performed via a Meta Quest 3s interface, mapping the operator's egocentric viewing frame directly to the robot's KingFisher head camera observation stream. For hardware safety and structural complexity reduction, the wheeled mobile base is locked, and the vertical torso elevation is pinned to a constant parameter. During both teleoperation and autonomous rollout evaluation, only the dual-arm chains and the central rotational waist joint remain active. The operational joint trajectories are continuously solved via an analytical Inverse Kinematics (IK) solver map based on end-effector poses. To guarantee continuous material handling without degradation during long-horizon evaluations, real-world granular corn kernels are substituted with standardized yellow foam balls. We collect about \textbf{1,200 highly dexterous demonstrations} with trajectory lengths \textit{spanning 45 to 65 seconds}. The task contains four compounded structural bottlenecks: 1) stable dexterous grasping of a highly compliant frustum-shaped paper cup, 2) picking up and orienting a rigid scooping shovel with the opposite hand, 3) executing a deep granular scoop to fill the shovel cavity, and 4) executing multi-limb spatial synchronization to pour the granular materials smoothly into the target cup without spilling.

\subsection{DexForce W1 Humanoid Setup: \texttt{Handover Book}}
\label{subsec:w1_handover}

\textbf{Hardware and Sensory Configuration:} To maximize grasping rigidity for rigid-object transfer, the W1 wheeled humanoid's end-effectors are swapped from dexterous hands to dual parallel-jaw PiPER grippers (Fig.~\ref{platforms}(c)). The camera allocation identical to the previous setup is maintained (KingFisher CV1 head camera and dual RealSense D405 wrist cameras running at 10 FPS), though the task structure functionally employs only the right robotic appendage for book manipulation.

\textbf{Task Definition and Constraints:} Data collection follows the same base-locked, torso-fixed Meta Quest 3s architecture. However, to capture human-interactive parameters, the motor controlling the vertical pitch axis of the neck is unfixed. The camera pitch velocity is directly linked via an API to the vertical height offset of the primary end-effector relative to the central base coordinate frame. This programmatic pairing guarantees that during the initial tabletop search phase, the neck actively pitches downward to maximize the field of view over the workspace, and subsequently pitches upward as the book is elevated to focus directly on the interacting human's hand. This removes the necessity of modifying the action expert architecture to predict extra neck joints. We log \textbf{500 interactive demonstrations} \textit{varying from 10 to 20 seconds}, where the timing variations result from randomized human hand trajectories or manual repositioning during the pre-grasping. The policy must satisfy a dual mandate: robust semantic instruction compliance (e.g., isolating and grasping a target blue versus brown book) and adaptive spatial agility when handling dynamic human-in-the-loop handovers across different human partners.

\subsection{DexForce A1 Semi-Humanoid Setup: \texttt{Fold Cloth}}
\label{subsec:a1_fold}

\textbf{Hardware and Sensory Configuration:} This task is validated on the DexForce A1 fixed-base semi-humanoid robot, which omits the liftable torso joint below the waist while retaining the upper dual-arm kinematic architecture and head modules (Fig.~\ref{platforms}(d)). To optimize edge-line fabric grasping, the robot uses parallel-jaw Pika grippers alongside the standard KingFisher CV1 head and dual RealSense D405 wrist camera suite configuration (all cameras are synchronized at 30 FPS).

\textbf{Task Definition and Constraints:} Operators drive the system using the Meta Quest 3s framework. Because the cloth manipulation space remains strictly bounded within the immediate frontal workspace, the head camera pitch angle is set to a static downward inclination during both data collection and real robot deployment phases. We accumulate a targeted dataset of \textbf{200 demonstrations}. The garment is initially placed at the geometric workspace center, alternating between pre-flattened states and highly unconstrained crumpled configurations. The operational trajectory duration \textit{scales from 60 to 90 seconds}. This duration is slightly longer than the corresponding Agilex mobile arm setup because driving a high-DoF semi-humanoid morphology via egocentric VR interfaces introduces higher teleoperation latency and execution overhead than direct master-slave mechanical tracking.

%% file: tex/suppModelDetails.tex
\section{More Details of Proposed Framework Dex-BEV}\label{appB}

\subsection{BEV Image Construction}

As proposed in Sec.~3.3 of the main text, we propose to synthesize BEV images from multi-view raw observations after designation of BEV frames. This BEV image is constructed by a \textit{top-down} orthographic projection of the aggregated colored point clouds from all cameras, where ``top-down'' is defined as the $z$-axis of the BEV frame. We select a 2D region-of-interest of 1.5 meter centered at the origin of the BEV frame, as the region to compute top-down orthographic projection. The color point clouds are rasterized into a (RGB) image with a fixed resolution at 224$\times$224. During the rasterization, we compute a height map that is pixel-wise aligned with the RGB BEV image. During network training, this height map is further converted into a vertex map (expressed in BEV frame), and feed into the policy similar to other RGB images and vertex maps (from raw camera views).

\subsection{Temporal Alignment in Trajectory Processing}

To eliminate the pseudo-motion noise caused by non-uniform execution speeds in human-teleoperated demonstrations~\cite{shi2026diversity}, we apply a velocity-based temporal normalization. For a given trajectory segment $\mathbf{A}_t\!=\!\{\mathbf{a}_t\}_{t=1}^K$, we compute the translational displacement $\Delta L_t\!=\!\| \mathbf{p}_{t+1} - \mathbf{p}_t \|_2$ and the rotational displacement $\Delta \theta_t\!=\!2 \arccos(|\langle \mathbf{q}_{t+1}, \mathbf{q}_t \rangle|)$, where $\mathbf{p}\!\in\!\mathbb{R}^3$ and $\mathbf{q}\!\in\!\mathcal{SO}(3)$ denote the position and quaternion orientation of the end-effector. The normalized time interval $\Delta \tau_t$ is determined by:
\begin{equation}
	\centering
	\Delta \tau_t = \max \left( \frac{\Delta L_t}{v_{std}}, \frac{\Delta \theta_t}{\omega_{std}}\right),
	\label{eqnE}
\end{equation}
\noindent where $v_{std}$ and $\omega_{std}$ are pre-defined standard velocities for quasi-static manipulation. For more than one robot arm, the $\Delta \tau_t$ is the maximum of two arms.
For movement that are almost static, i.e. $\Delta \tau_t \approx 0$, we would either use the original duration or directly drop this frame (if the static ``waiting'' is not relevant to the manipulation task).
During training, we perform cubic spline interpolation to obtain the action chunk.

%% file: tex/suppExpResults.tex
\section{More Details of Various Experimental Results}\label{appC}

This section provides extended quantitative data from our simulation benchmarks and provides comprehensive qualitative analyses, case studies, and keyframe breakdowns from our real-world evaluations across multiple robotic embodiments. Comprehensive full-length rollouts for all real-world tasks are provided in the \textbf{Supplementary Videos}.

\begin{figure}[h]
	\begin{center}
    \includegraphics[width=1.0\linewidth]{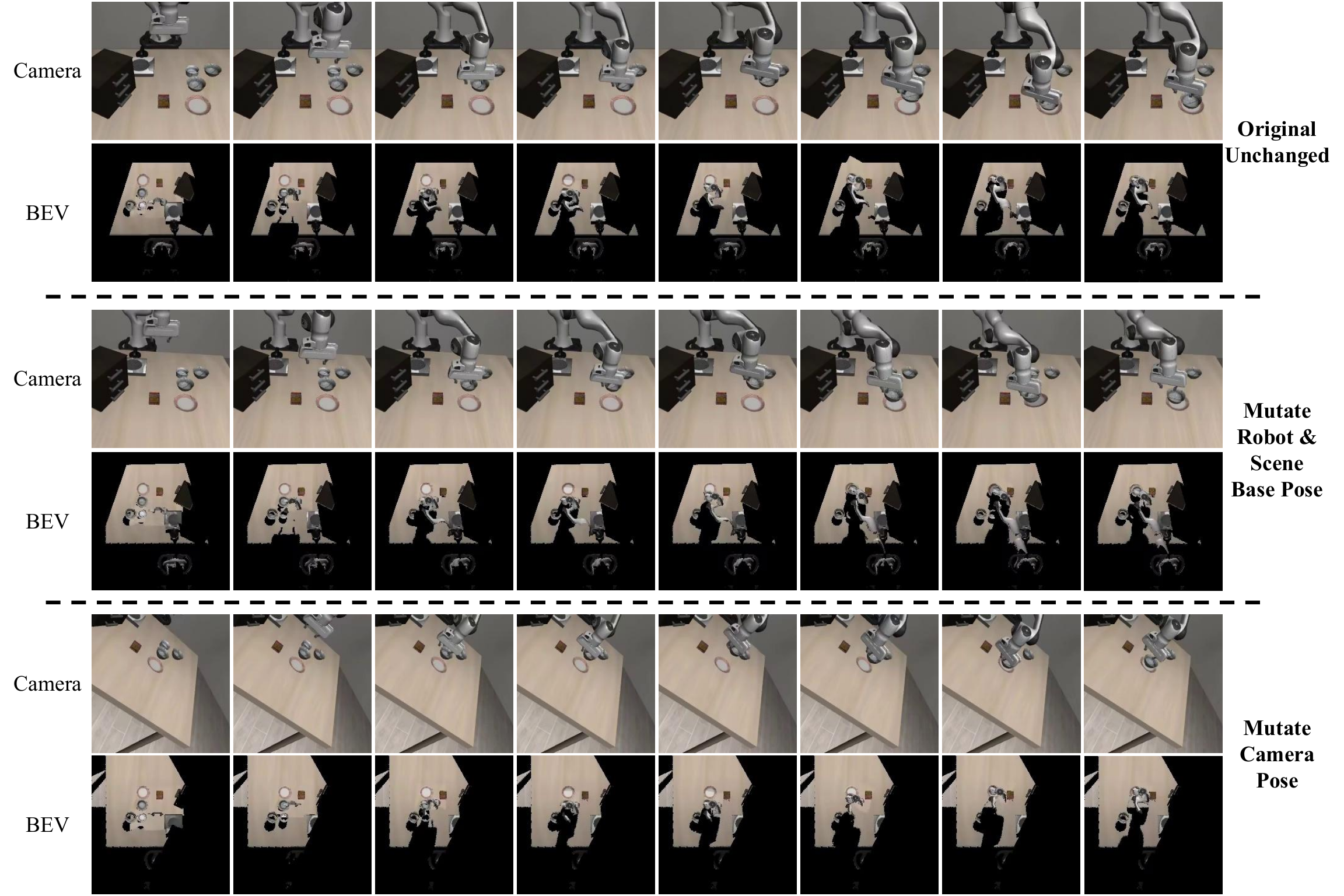}
	\vspace{-15pt}
	\caption{\textbf{Examples of modified LIBERO.} From top to bottom, these represent observations with no parameters modified, observations with modifications to the robot and workstation, and observations with only the camera pose modified, respectively.} 
    \label{modifiedLIBERO}
	\vspace{-10pt}
	\end{center}
\end{figure}

\subsection{Simulation Benchmarks and Ablation Studies}
\label{subsec:sim_detailed_results}

For the LIBERO~\cite{liu2023libero} and RoboTwin 2.0~\cite{mu2025robotwin} benchmarks, we first use the official setups regarding robots, training data and evaluation protocol. On the other hand, we modify the camera pose, robot base pose and scene pose to evaluate the robustness of our proposed method Dex-BEV with respect to these perturbations, as detailed below (some modified examples can be found in Fig.~\ref{modifiedLIBERO}).

For the camera pose, we apply the following randomization on translation and rotation. Given the canonical pose of the third-person view camera, we apply rotation with respect to the world $z$-axis, rotation to change the angle of tilt, and rotation with respect to the optical axis of the camera. We used uniform distributions of the rotation angles among each case, with ranges of 140/60/60 degrees for each axis. Moreover, we randomize the distance from the camera to the center of the scene, by uniformly sampling from an interval centered at the canonical distance. The range of the interval is 1 meter. For each trajectory rollout, the camera pose is randomly reset at the beginning and kept static. To ensure that the relevant objects have sufficient visibility, we filter the sampled camera pose using the following criterion: the number of points (computed from depth image and camera parameters) in a 3D region-of-interest (manually selected for each scene) must exceed a given threshold.

After permuting the camera pose, we apply a randomly permutation of the base pose of the robot and \textit{scene}, where scene implies every item except the robot (usually including objects on the table). At the beginning of the trajectory rollout, we apply a translational permutation of 10 cm and a rotational permutation of 5 degrees on all $x$, $y$, and $z$ axes. After the permutation, we move the robot end-effector pose to compensate for the offset caused by robot and scene movements. The official demonstration trajectories are re-used by applying similar compensations. We would filter out randomly sampled robot and scene base pose pair by kinematic reachability. 

From Tab.~2 of the main text, directly evaluating VLA trained with trajectories in official setups, such as official X-VLA checkpoints~\cite{zheng2026xvla}, leads to a nearly zero success rate. Moreover, the 2D ablation baselines, even trained with mutated setups, cannot absorb the pose variation of the training data and yield the success rate again nearly zero. In comparison, the proposed method achieves reasonable success rate with aligned 3D input \& output and view-invariant BEV images. To provide a clearer view of the specific performance of various previous VLA approaches, we present a more comprehensive comparison with multiple prior methods in Tab.~\ref{tabSimResultsMixedAppd}.

\begin{table}[]
    \centering
    \caption{\textbf{Simulation benchmark results and generalization to different embodiments.} We present the success rate for each compared method across task suites in LIBERO~\cite{liu2023libero} and RoboTwin 2.0~\cite{mu2025robotwin}. This table is the complete supplementary version of Tab.~1 in the main text.
    }
	\setlength{\tabcolsep}{1pt}
    \begin{tabular}{c|c|cccc|c|cc}
	\Xhline{1.2pt}
    \multirow{2}{*}{Method} & \multirow{2}{*}{\begin{tabular}[c]{@{}c@{}}Cross \\ Embodiments\end{tabular}} & \multicolumn{5}{c|}{LIBERO (Official)}    & \multicolumn{2}{c}{RoboTwin 2.0} \\
    \cline{3-9} 
    ~ & ~ & \cellcolor{gray!15} Spatial & \cellcolor{gray!15} Object & \cellcolor{gray!15} Goal & \cellcolor{gray!15} Long & \cellcolor{gray!15} Average & \cellcolor{gray!15} Clean & \cellcolor{gray!15} Randomized        \\
    \hline
    DP3~\cite{ze20243d}             & False & --- & --- & --- & --- & --- & 55.2 & 5.0 \\
    RDT-1B~\cite{liu2025rdt}        & False & --- & --- & --- & --- & --- & 34.5 & 13.7 \\
	TraceVLA~\cite{zheng2025tracevla} & False & 84.6 & 85.2 & 75.1 & 54.1 & 74.8 & --- & --- \\
	Octo~\cite{team2024octo} & False & 78.9 & 85.7 & 84.6 & 51.1 & 75.1 & --- & --- \\
	OpenVLA~\cite{kim2025openvla} & False & 84.7 & 88.4 & 79.2 & 53.7 & 76.5 & --- & --- \\
	SpatialVLA~\cite{qu2025spatialvla} & False & 88.2 & 89.9 & 78.6 & 55.5 & 78.1 & --- & --- \\
	4D-VLA~\cite{zhang20254d} & False & 88.9 & 95.2 & 90.9 & 79.1 & 88.6 & --- & --- \\ 
	DreamVLA~\cite{zhang2025dreamvla} & False & 97.5 & 94.0 & 89.5 & 89.5 & 92.6 & --- & --- \\ 
    $\pi_0$~\cite{black2025pi0}     & False & 96.8 & 98.8 & 95.8 & 85.2 & 94.2 & 46.4 & 16.4 \\
	DepthVLA~\cite{yuan2025depthvla} & False & 96.4 & 98.0 & 95.8 & 89.2 & 94.9 & --- & --- \\  
	UniVLA~\cite{bu2025learning} & False & 95.4 & 98.8 & 93.6 & 94.0 & 95.4 & --- & --- \\
	GeoPredict~\cite{qian2025geopredict} & False & 98.0 & 98.2 & 95.7 & 94.0 & 96.5 & --- & --- \\ 
	OpenVLA-OFT~\cite{kim2025fine} & False & 97.6 & 98.4 & 97.9 & 94.5 & 97.1 & --- & --- \\
	GeoVLA~\cite{sun2025geovla} & False & 98.4 & 99.0 & 96.6 & 96.6 & 97.7 & --- & --- \\ 
    X-VLA~\cite{zheng2026xvla}      & False & 98.2 & 98.6 & 97.8 & 97.6 & 98.1 & 70.0 & 39.0 \\
    \hline
    2D Ablation                     & True  & 93.2 & 95.0 & 92.8 & 90.2 & 92.8 & 64.8 & 35.2 \\
    Dex-BEV                        & True  & 98.2 & 98.0 & 97.8 & 97.0 & 97.8 & 76.0 & 42.0 \\
	\Xhline{1.2pt}
    \end{tabular}
    \label{tabSimResultsMixedAppd}
    \vspace{-10pt}
\end{table}

\subsection{Agilex Bimanual Evaluations: \texttt{Fold Mailer Box \& Fold Cloth}}
\label{subsec:agilex_detailed}

\begin{figure}[h]
	\begin{center}
    \includegraphics[width=1.0\linewidth]{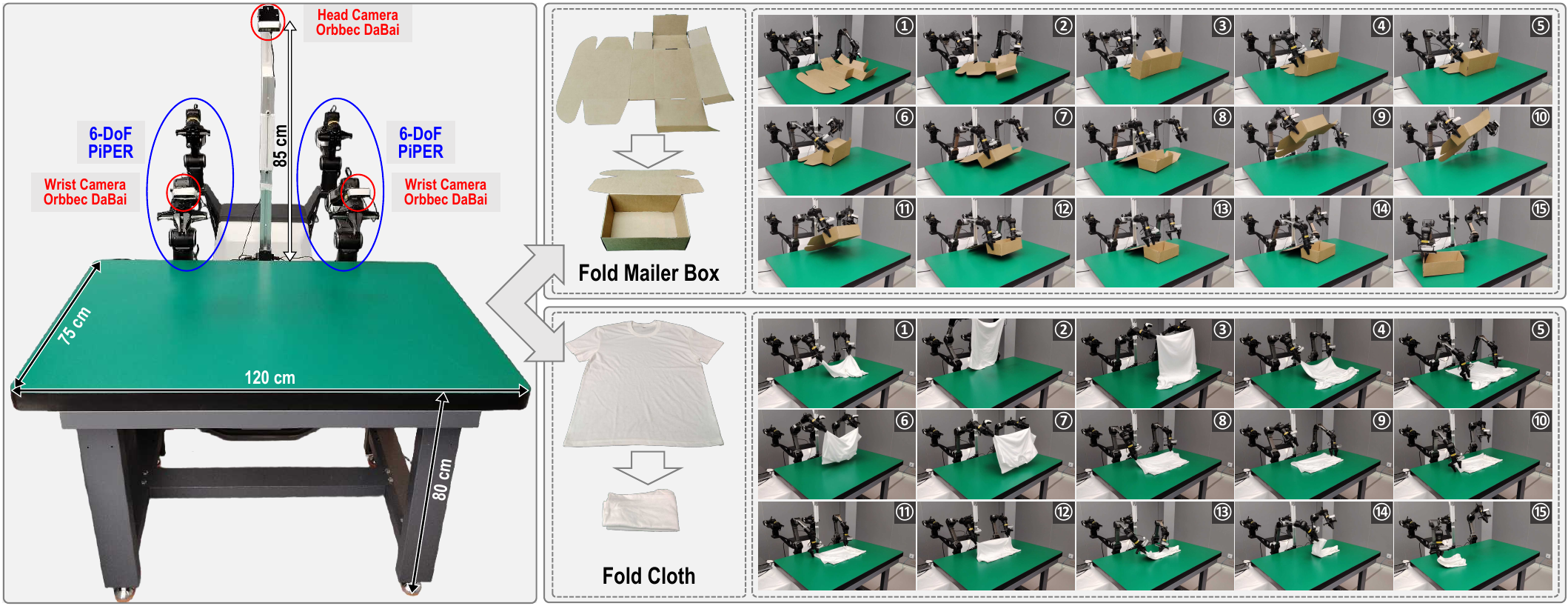}
	\vspace{-15pt}
	\caption{\textbf{Real-world bimanual platform and task execution.} (\textit{Left}) Configuration of the Agilex bimanual robotic platform used for data collection and inference. (\textit{Right}) Detailed view of target objects and sequential keyframes from autonomous rollouts of two challenging long-horizon tasks: \texttt{Fold Mailer Box} (articulated) and \texttt{Fold Cloth} (deformable). These snapshots illustrate Dex-BEV's capability in handling complex spatial reasoning and multi-arm coordination. } 
    \label{platforms1}
	\vspace{-10pt}
	\end{center}
\end{figure}

\textbf{In-Distribution Keyframe Rollouts:}
Fig.~\ref{platforms1} illustrates the chronological keyframe sequences of autonomous in-distribution (ID) executions for tasks \texttt{Fold Mailer Box} and \texttt{Fold Cloth} on the Agilex bimanual platform. These long-horizon tasks represent some of the most intricate dexterous manipulation challenges in current literature \cite{black2025pi0, zheng2026xvla}. Notably, while state-of-the-art models like $\pi_0$ require approximately 1,000 hours of garment data to achieve policy convergence, and X-VLA limits this requirement to roughly 1,500 demonstrations ($\sim$25 hours), Dex-BEV accomplishes a higher average success rate using only around 400 demonstrations. This reduction to less than one-third of X-VLA's data footprint highlights our framework's superior data efficiency and its readiness for rapid on-site deployment. Similarly, for the complex \texttt{Fold Mailer Box} task, the 1,500 fine-tuning demonstrations translate to only about 17 total hours of execution time—significantly lower than data requirements for comparable dexterous skills in literature \cite{li2026causal, team2026gigabrain}.

\textbf{Out-of-Distribution (OOD) and Error Recovery Tests:}
To evaluate policy resilience, we introduce various rigorous OOD perturbations.

\begin{figure}[h]
	\begin{center}
    \includegraphics[width=1.0\linewidth]{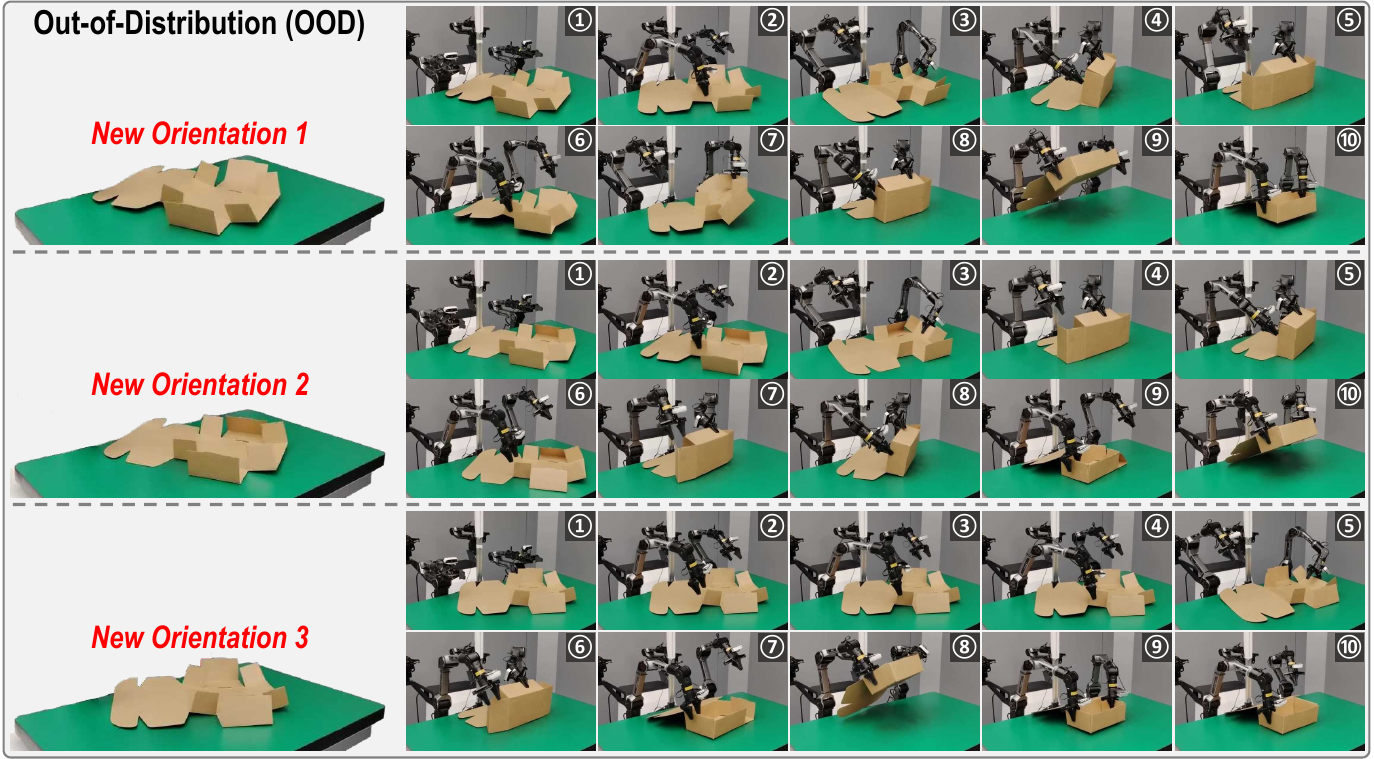}
	\vspace{-15pt}
	\caption{\textbf{Rollouts of Self-Recovery and Orientation Invariance.} For task \texttt{Fold Mailer Box}, we demonstrate the qualitative results under the OOD trails. It is best to zoom in to view the details. } 
    \label{platforms1-OOD1}
	\vspace{-10pt}
	\end{center}
\end{figure}

\textbf{\textit{(1) Self-Recovery and Orientation Invariance:}} For the task \texttt{Fold Mailer Box}, the box is initialized with unseen poses and extreme yaw angles. As shown in Fig.~\ref{platforms1-OOD1}, Dext-BEV utilizes closed-loop visual servoing to execute pre-manipulation re-orientation steps, aligning the box before initiating the folding sequence. This spatial awareness enables robust error self-recovery. If a box slips mid-execution, the policy autonomously recovers from the anomalous state without human intervention.

\textbf{\textit{(2) Continuous Operation Facility:}} Still for the task \texttt{Fold Mailer Box}, we demonstrate the policy's capacity for continuous, multi-cycle operation in the \textbf{Supplementary Videos}. After completing a box, the right arm clears the workspace, both arms return to their home configurations, and a new box blueprint is immediately introduced. Backed by high single-cycle success rates, Dex-BEV reliably handles 3 to 5 continuous, un-interrupted folding rollouts.

\begin{figure}[h]
	\begin{center}
    \includegraphics[width=1.0\linewidth]{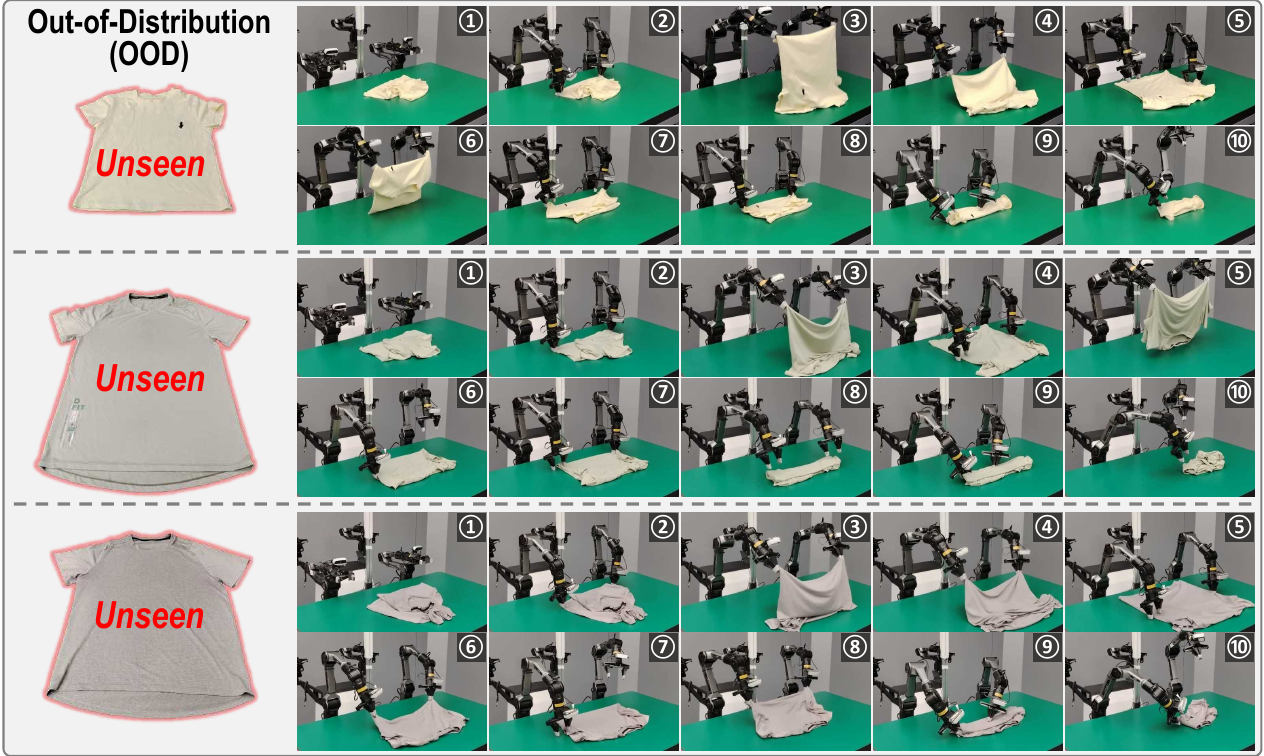}
	\vspace{-15pt}
	\caption{\textbf{Rollouts of Zero-Shot Instance Generalization.} For task \texttt{Fold Cloth}, we demonstrate the qualitative results under the OOD trails. It is best to zoom in to view the details.} 
    \label{platforms1-OOD2}
	\vspace{-10pt}
	\end{center}
\end{figure}

\textbf{\textit{(3) Zero-Shot Instance Generalization:}} For the task \texttt{Fold Cloth}, the model is trained exclusively on white XL/XXL T-shirts. In OOD trials, we evaluate the system against a small beige S-sized shirt, a light green XXL shirt, and a gray XXL shirt. Fig.~\ref{platforms1-OOD2} validates that the model generalizes zero-shot across diverse colors, geometries, and scales. 

\subsection{DexForce W1 Humanoid Evaluations: \texttt{Scoop Popcorn}}
\label{subsec:w1_scoop_detailed}

\begin{figure}[h]
	\begin{center}
    \includegraphics[width=1.0\linewidth]{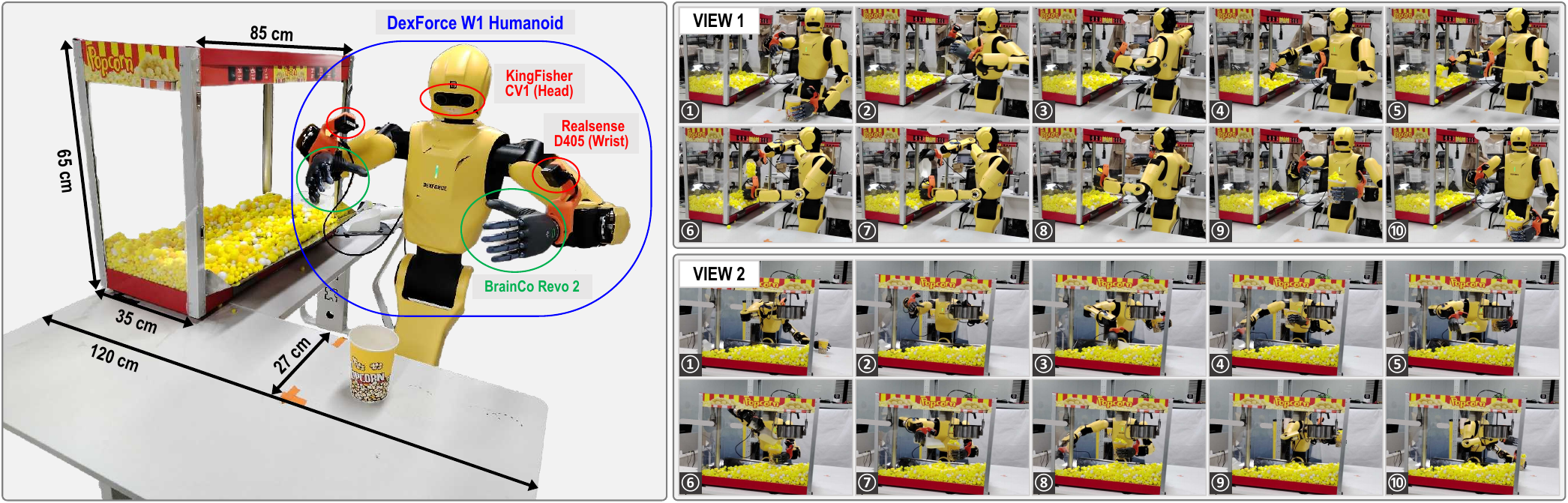}
	\vspace{-15pt}
	\caption{\textbf{Bimanual humanoid platform and long-horizon task execution.} (\textit{Left}) Detailed configuration of the DexForce W1 humanoid robot with two dexterous hands and its operation environment. (\textit{Right}) Multi-view keyframes showcasing the autonomous rollout of the \texttt{Scoop Popcorn} task. This complex, long-horizon sequence requires fine-grained bimanual coordination to manipulate the paper cup while simultaneously scooping and filling it with a shovel. } 
    \label{platforms2}
	\vspace{-10pt}
	\end{center}
\end{figure}

\textbf{Multi-View Keyframe Breakdowns:} Fig.~\ref{platforms2} presents the chronological execution of the \texttt{Scoop Popcorn} task from two complementary viewing angles. This task couples tool-use, high-DoF anthropomorphic dexterity, dense multi-object contact, granular material estimation, and wide-range trajectory tracking. While similar granular tasks have been demonstrated on humanoid hardware platforms like Tesla Optimus, technical details are omitted in public literature, and autonomous execution validity remains unverified]. In contrast, Dex-BEV handles this task autonomously with a compact post-training dataset of approximately 18 total hours, providing a highly scalable receipt for rapid deployment in new manipulation scenarios.

\begin{figure}[h]
	\begin{center}
    \includegraphics[width=1.0\linewidth]{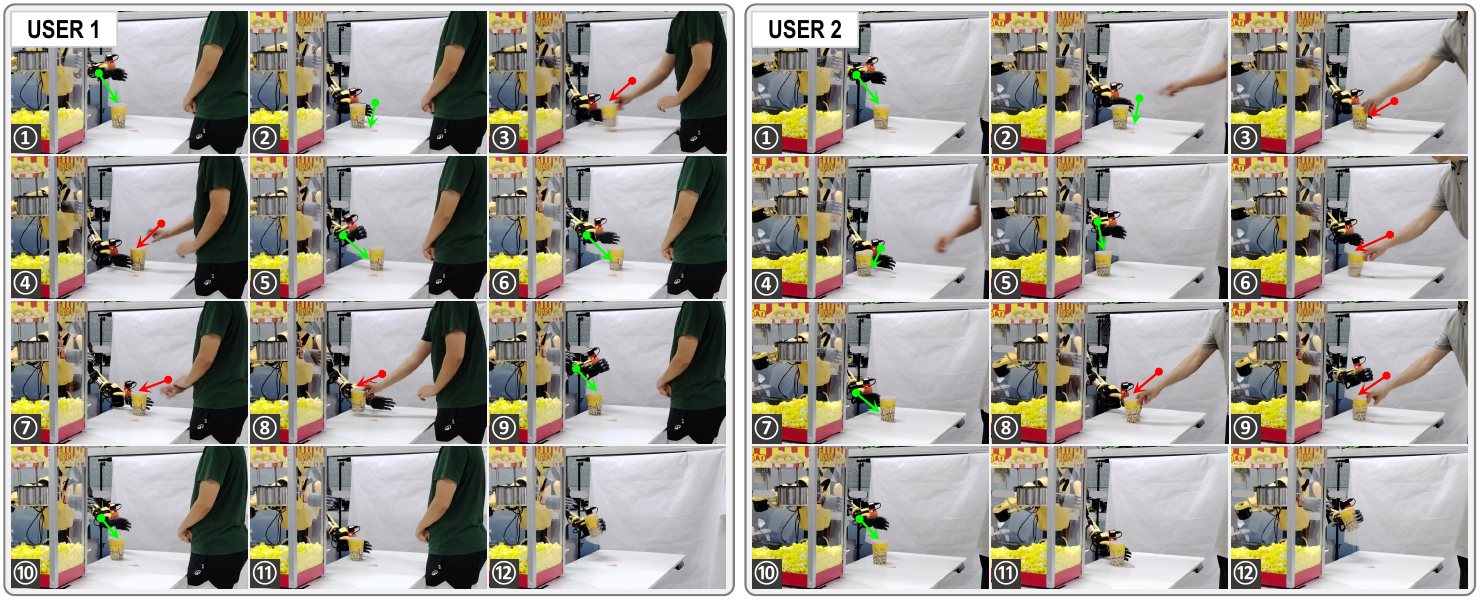}
	\vspace{-15pt}
	\caption{\textbf{Robustness to dynamic interference.} Sequential snapshots from the cup-grasping phase of the \texttt{popcorn scooping} task on the DexForce W1 platform. The images demonstrate the model’s real-time reactivity. Despite random manual displacements of the target cup by two different users, Dex-BEV successfully recalibrates the motion trajectory to achieve a successful grasp. This highlights the closed-loop robustness of the proposed framework against external disturbances. } 
    \label{platforms2-OOD}
	\vspace{-10pt}
	\end{center}
\end{figure}

\textbf{Dynamic Adversarial Robustness:} Although adversarial samples or shifting targets were entirely absent during data collection, we introduce active human intervention during the OOD testing phase. During the robot's pre-grasping approach, multiple human operators dynamically and repeatedly shift the target paper cup's location. As verified in Fig.~\ref{platforms2-OOD}, Dex-BEV dynamically perceives the cup’s displacement, smoothly retracts its arms, recalculates its spatial trajectory, and successfully reseals the grasp. This resistance to un-modeled workspace disturbances underscores the reactivity enabled by our unified 3D BEV observation space.

\begin{figure}[h]
	\begin{center}
    \includegraphics[width=1.0\linewidth]{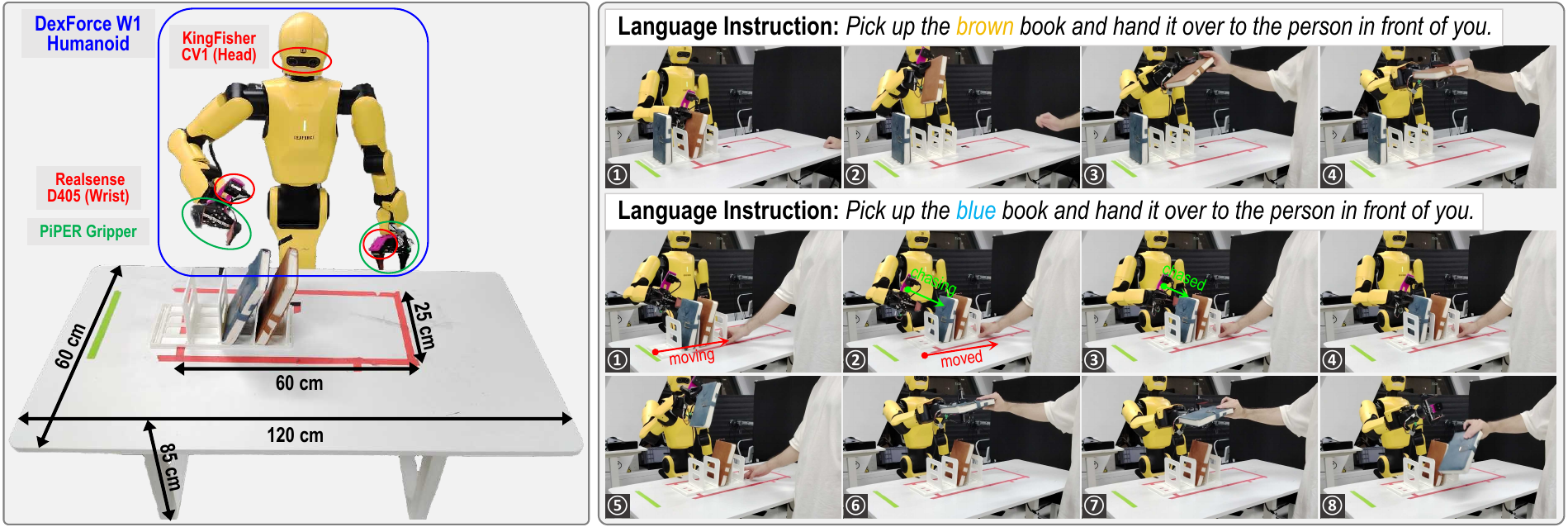}
	\vspace{-15pt}
	\caption{\textbf{Multi-modal interactive handover book task.} (\textit{Left}) DexForce W1 humanoid platform with two grippers and its workspace setup. (\textit{Right}) Successive keyframes of the robot executing a \texttt{Handover Book} task conditioned on different language instructions. The sequences highlight Dex-BEV’s ability to interpret color-specific commands and perform precise, interactive maneuvers for handing over target objects to a human partner. } 
    \label{platforms3}
	\vspace{-10pt}
	\end{center}
\end{figure}

\subsection{DexForce W1 Humanoid Evaluations: \texttt{Handover Book}}
\label{subsec:w1_handover_detailed}

Fig.~\ref{platforms3} details the interactive task \texttt{Handover Book} evaluation under diverse language instructions and dynamic workspace shifting, showing Dex-BEV's superiority on semantic grounding and interactive tracking. The task also verifies the model's semantic sensitivity. The policy isolates and tracks specific targets based on user-specified attributes (e.g., fetching a \textit{blue} versus a \textit{brown} book). Furthermore, during the grasping phase, an operator actively shifts and rotates the underlying bookshelf (indicated by red arrows). As shown by the tracking vectors (indicated by green arrows), the arm recalculates its relative trajectory in real-time to complete the grasp. Once the object is elevated and moved toward the user, the policy tracks the user's hand and maintains its grasp until it senses firm physical contact and a steady receipt by the human partner, at which point it opens the parallel jaws and safely returns to its home configuration.

\subsection{DexForce A1 Semi-Humanoid Evaluations: \texttt{Fold Cloth}}
\label{subsec:a1_fold_detailed}

\begin{figure}[h]
	\begin{center}
    \includegraphics[width=1.0\linewidth]{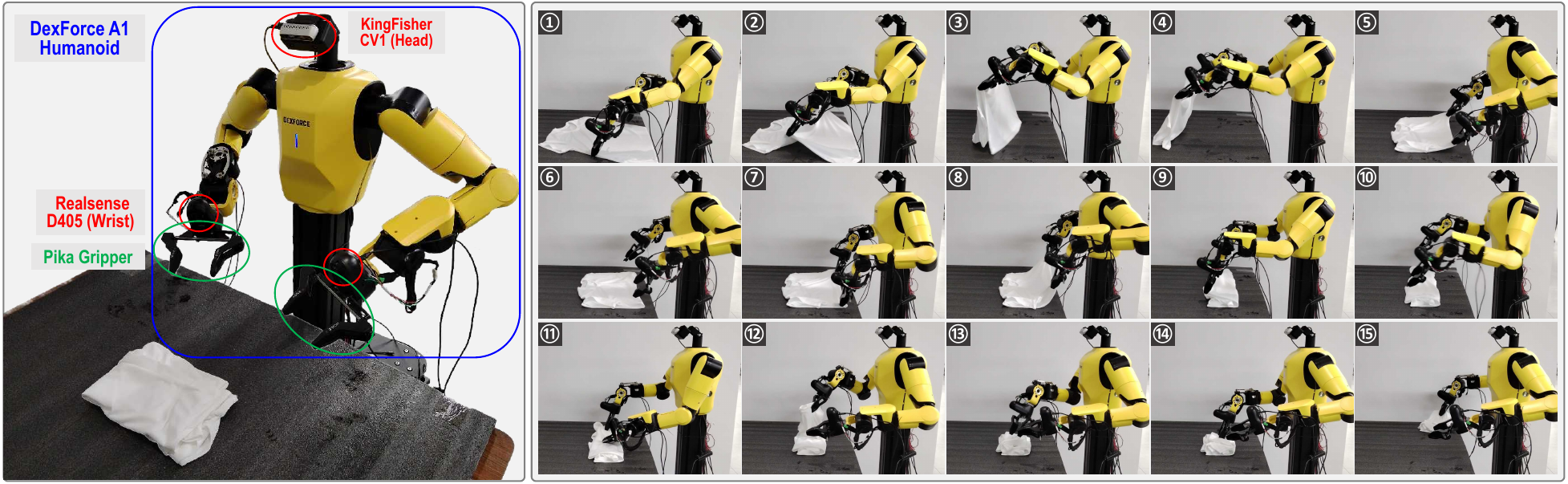}
	\vspace{-15pt}
	\caption{\textbf{Bimanual garment folding on the DexForce A1 platform.} (\textit{Left}) Overview of the semi-humanoid hardware configuration and experimental workspace for the task \texttt{Fold Cloth}. (\textit{Right}) Sequential keyframes of an autonomous rollout demonstrating long-horizon manipulation of a deformable garment. The sequence highlights the framework’s ability to coordinate dual-arm trajectories for complex fabric manipulation while the robot’s lower body remains stationary. } 
    \label{platforms4}
	\vspace{-10pt}
	\end{center}
\end{figure}

Fig.~\ref{platforms4} documents the cloth folding task executed on the A1 semi-humanoid platform, showing Dex-BEV's capability about the distinct state segmentation and human-like rollout trajectories. The policy successfully handles two distinct structural initialization states: a flat canonical layout and an unconstrained, crumpled pile. Dex-BEV accurately segments the task phases, flattening the crumpled garment prior to executing the folding sequence. This adaptation uses only 200 demonstration trajectories—fewer than the 400 demonstrations used on the Agilex dual-arm platform. This efficiency stems from a more constrained frontal workspace layout that reduces the necessary spatial sampling density.
Interestingly, due to its anthropomorphic shoulder configuration and elevated workspace clearance, the A1 platform generates more human-like arm trajectories compared to the table-bound Agilex arm setup. The greater vertical range allows the A1 robot to lift, smooth, and align the fabric layers with high precision, yielding flatter, wrinkle-free folds. These qualitative differences highlight the impact of data diversity and embodiment kinematics on downstream policy behavioral traits.

%% file: tex/suppDiscussions.tex
\section{More Discussions of Limitations and Future Works}\label{appD}

To provide a critical and transparent evaluation of the Dex-BEV framework, this section unpacks the systemic limitations of our current methodology, analyzes the corner cases and failure modes observed across our five real-world benchmarks, and outlines strategic pathways for future enhancements in embodied foundation models.

\subsection{Methodological Limitations of Dex-BEV}
\label{subsec:method_limitations}

While Dex-BEV successfully establishes a unified 3D coordinate system for multi-view observations and actions, it inherits a strong dependency on precise camera calibration, which is relatively easy to obtain in simulation. The structural integrity of the synthesized Bird's-Eye-View (BEV) images and vertex maps relies heavily on the accuracy of the extrinsic matrices $\mathbf{T}_{t,i}$. In real-world deployments, subtle hardware vibrations, thermal expansion of robot links, or accidental physical contact can introduce extrinsic drift, leading to geometric distortion or pixel-to-vertex misalignment in the BEV projection. Furthermore, our current depth relaxation strategy (Vertex Spectrum) via linear-increasing discretization (LID) samples a fixed number of depth hypotheses. While computationally efficient, this deterministic quantization introduces spatial discretization errors in fine-grained dexterous zones, occasionally rounding off the sub-centimeter geometric boundaries required for tight multi-finger interactions. Lastly, the current pipeline processes historical data over a relatively short temporal window, which restricts the policy's capacity to build abstract semantic representations of long-horizon task progress independent of immediate geometric changes.

\subsection{Failure Mode Analysis of Real-World Benchmarks}
\label{subsec:real_world_failures}

Despite achieving state-of-the-art success rates across diverse dual-arm manipulation platforms, our real-world evaluations did not achieve a 100\% success rate due to a combination of hardware constraints, kinematic limits, and un-modeled environmental stochasticity:

\begin{itemize}[leftmargin=2em]
    \item \textit{Hardware Instability and Mechanical Fatigue:} Under long-duration un-interrupted testing, mechanical wear introduces backlash in the high-DoF anthropomorphic hands and parallel grippers. This physical degradation reduces joint tracking accuracy, resulting in micro-slips during the grasp phase of the \texttt{Fold Mailer Box} and \texttt{Scoop Popcorn} tasks.
    \item \textit{Kinematic Reachability and Fixed-Base Constraints:} Because the bases of the Agilex, A1 and W1 humanoids are locked during our experiments to ensure safety, the dual-arm systems occasionally encounter kinematic singularities or reachability limits. For example, if a cloth or a mailer box is randomly placed near the extreme boundary of the workspace during an out-of-distribution (OOD) trial, the optimal end-effector trajectory derived by the policy cannot be executed due to joint-space limits, causing the sub-step to fail.
    \item \textit{Unseen Geometric and Material OOD Distributions:} While the policy exhibits impressive zero-shot generalization across colors and scales, extreme variations in object material properties remain a bottleneck. In the \texttt{Fold Cloth} task, deploying a garment with highly specular, silky fabric or extreme stiffness causes errors in both the pre-trained VLM's feature extraction and the geometric projection, resulting in anomalous folding actions.
    \item \textit{Perceptual Distortions from Extreme Environmental Shifts:} Drastic scene-level lighting variations, heavy shadows cast by human operators during the \texttt{Handover Book} task, or highly reflective tabletop backgrounds degrade the quality of the multi-view visual inputs. These severe visual perturbations propagate through the large VLM backbone, producing jittery action sequences or premature jaw releases.
\end{itemize}

\subsection{Future Research Directions}
\label{subsec:future_works}

To address these challenges and maximize the real-world utility of 3D-aligned policy learning, future developments will focus on two major dimensions:

\textbf{(1) Algorithmic and Data Infrastructure Enhancements:}
\begin{itemize}[leftmargin=2em]
    \item \textit{World-Action Model (WAM) Integration:} Expanding Dex-BEV into a generative 3D World-Action Model will enable the policy to predict future 3D BEV states and point cloud rollouts concurrently with action generation. This forward-prediction capability will allow the robot to perform mental rollouts and self-correct trajectories before execution.
    \item \textit{Lightweight and Long-Horizon Memorable VLAs:} We aim to compress the VLM backbone into a specialized, high-frequency edge-VLA model to lower inference latency on consumer hardware. Simultaneously, integrating state-space models (e.g., Mamba architectures) or advanced transformer sequence-modeling techniques will equip the VLA with long-range memory, allowing it to maintain abstract task context over hundreds of steps.
    \item \textit{Leveraging 3D Vision Foundation Models for Scaling}: We plan to incorporate rapidly maturing 3D vision foundation models—including advanced SLAM frameworks, multi-view geometry reconstruction pipelines, and novel view or scene synthesis models—directly into our data preparation infrastructure. By utilizing these models to pre-process large-scale robotic manipulation datasets and egocentric human demonstration videos, we can automatically extract precise camera trajectories and dense spatial annotations. This paradigm will drastically lower the cost of generating highly accurate, geometrically consistent, and diverse 3D-aligned pre-training data at scale, further amplifying the framework's cross-embodiment generalization.
\end{itemize}

\textbf{(2) Hardware Capabilities and Multi-Modal Grounding:}
\begin{itemize}[leftmargin=2em]
    \item \textit{Transition to Mobile Manipulation:} Unlocking the full wheel-foot locomotion capabilities of the Agilex, W1 and A1 platforms will transition the framework from tabletop setups to unconstrained, room-scale mobile manipulation tasks.
    \item \textit{Multi-Agent Robotic Collaboration:} Extending our unified BEV coordinate mapping to distributed, multi-robot systems will allow multiple distinct embodiments to share a single canonical spatial observation grid, enabling zero-shot multi-agent cooperative manipulation.
    \item \textit{Multi-Modal Sensory Fusion:} To complement visual perception, we plan to ingest high-frequency force-tactile, and auditory feedback arrays into the VLA architecture. Incorporating tactile streaming will resolve the visual occlusions that occur during tight bimanual handovers, establishing a truly robust, multi-modal interface for generalizable embodied intelligence.
\end{itemize}